%% file: iccv_se3_main.tex
\documentclass{article}

\usepackage{arxiv}
\usepackage{times}
\usepackage{epsfig}
\usepackage{graphicx}
\usepackage{amsmath}
\usepackage{amssymb}
\usepackage{bbm}
\usepackage{color}
\usepackage{algorithm}
\usepackage{algorithmicx}
\usepackage{algpseudocode}
\usepackage{hhline}
\usepackage{caption,subcaption}
\usepackage{tabularx}
\usepackage{ulem}
\usepackage{rotating}
\usepackage{array}
\usepackage{multirow,comment}
\input{myincludes.tex}

\newcolumntype{L}[1]{>{\raggedright\let\newline\\\arraybackslash\hspace{0pt}}m{#1}}
\newcolumntype{C}[1]{>{\centering\let\newline\\\arraybackslash\hspace{0pt}}m{#1}}
\newcolumntype{R}[1]{>{\raggedleft\let\newline\\\arraybackslash\hspace{0pt}}m{#1}}

\usepackage[pagebackref=true,breaklinks=true,letterpaper=true,colorlinks,bookmarks=false]{hyperref}

\begin{document}

\title{Efficient and Robust Registration on the 3D Special Euclidean Group}

\author{
  Uttaran Bhattacharya\\
  Department of Computer Science\\
  University of Maryland College Park\\
  \texttt{uttaranb@cs.umd.edu} \\
   \And
 Venu Madhav Govindu \\
  Department of Electrical Engineering\\
  Indian Institute of Science\\
  \texttt{venug@iisc.ac.in} \\
}

\maketitle

\begin{abstract}

We present an accurate, robust and fast method for registration of 3D scans. Our motion estimation optimizes a robust cost function on the intrinsic representation of rigid motions, \textit{i.e.}, the Special Euclidean group $\mathbb{SE}(3)$. We exploit the geometric properties of Lie groups as well as the robustness afforded by an iteratively reweighted least squares optimization. We also generalize our approach to a joint multiview method that simultaneously solves for the registration of a set of scans. We demonstrate the efficacy of our approach by thorough experimental validation. Our approach significantly outperforms the state-of-the-art robust 3D registration method based on a line process in terms of both speed and accuracy. We also show that this line process method is a special case of our principled geometric solution. Finally, we also present scenarios where global registration based on feature correspondences fails but multiview ICP based on our robust motion estimation is successful.

\end{abstract}

\section{Introduction}
The availability of consumer depth cameras has allowed us to acquire reliable 3D scans of a scene or an object~\cite{KinectFusion,sturm2012benchmark,Zhang2012}. To build a geometrically consistent 3D reconstruction, we need to solve the key problem of aligning or registering all scans in a global frame of reference~\cite{Salvi2007,Tam2013,Guo2014,Mellado2014}. The resulting solution can be used in a diverse range of contexts including human-computer interaction, archiving of cultural objects, industrial inspection etc. There is a wide variety of solutions to the 3D registration problem in the literature~\cite{Gelfand2005,Makadia2006,Holz2015,RRIS,FGR}. All of these methods involve a trade-off between speed and accuracy. Recently,~\cite{FGR} has presented a method for fast global registration (henceforth denoted FGR) of 3D scans based on the robustness of a line process. This method has been shown to outperform existing methods in terms of both speed and accuracy. \\
\indent As in~\cite{FGR}, given a set of 3D feature correspondences, we pose the registration problem as one of solving for the rigid Euclidean motion that minimizes a robust cost function. However, unlike their approach, our solution systematically utilises the rich geometric structure of the underlying Lie group representation for 3D motion, \textit{i.e.}, the Special Euclidean group $\mathbb{SE}(3)$. In this context, we achieve robustness via the iteratively reweighted least squares (IRLS) method. The key observation of our paper is that our specific combination of geometry of rigid Euclidean motion and the robustness of IRLS allows our method to significantly outperform the fast global registration method of~\cite{FGR} without having to take recourse to tuning of parameters. Additionally, we show that the solution proposed in~\cite{FGR} is a special case of our more general (and more accurate and faster) solution. In the process, we demonstrate that we can gain both theoretical insight as well as improved performance by utilizing the rich geometric structure of the underlying geometric representation of rigid Euclidean motions.\\
\indent Furthermore, our work also addresses two important considerations. Firstly, in order to achieve robustness, some loss functions used for optimization have parameters that need to be either provided \textit{a priori} or estimated \textit{in situ}. This is the case with~\cite{FGR} which uses the Geman-McClure loss function. We argue that for the problem of 3D registration the statistical efficiency (i.e. accuracy) trade-off inherent to all robust estimators can be easily optimized using loss functions that are parameter free. This obviates the need to estimate any parameter during the course of optimization. Secondly, we argue that while fast and accurate 3D registration can be achieved using 3D point feature correspondences, this approach has certain limitations. While such feature correspondences can be reliably obtained when the camera motion is small (equivalently there is significant overlap between scans), there are certain scenarios where the feature correspondences break down. For such cases, we demonstrate that accurate joint registration of multiple 3D scans can be achieved by incorporating our robust motion estimation method into local methods such as the ICP.

\section{Literature Survey}
Although the literature of 3D registration is extensive, in the following we only focus on aspects that are directly relevant to our method. A large number of methods for registering 3D scans using point correspondences have two key aspects, (a) a method for establishing point correspondences between two 3D scans and (b) a method for estimating motion given an adequate number of such correspondences. We may further classify methods according to whether they use a fixed set of point correspondences~\cite{RRIS,FGR,Tombari2013,Guo2016} or update them~\cite{Rusinkiewicz2001,Aiger2008,Drost2010,Mellado2014,Bouaziz}. In the former instance, we use invariant feature representations to find point correspondences across a pair of scans which are then used for motion estimation. In the latter approach, we alternately update point correspondences using nearest neighbors and motion estimation till convergence, such as in the classical approach of ICP and its variants (see~\cite{Rusinkiewicz2001,Bouaziz,chetverikov2002trimmed,yang2013go} and references therein). Independent of the method for establishing correspondences, we need a method for robust motion estimation method given a set of correspondences. This problem of robust motion estimation is the main focus of this paper.\\
\indent The least squares solution to the motion estimation problem is the classical method of Umeyama~\cite{Umeyama1991}. However, the least squares solution breaks down in the presence of outliers. The requisite robustness is often achieved using variants of RANSAC~\cite{Aiger2008,Raguram2008,FPFH,Papazov2012,Mellado2014,Holz2015} or motion clustering~\cite{Drost2010,Mellado2014,Salas2013}. Other classes of approaches are based on the branch-and-bound framework~\cite{Gelfand2005,Hartley2007,Li2007,Enqvist2009,Yang2016}. However, all of these approaches often require expensive iterations and are slow to converge. Other solutions based on expectation-maximization using Gaussian mixture model representations of the scans~\cite{JianVemuri,EvangelidisHoraud,EckartEtAl} are similarly slow. Another approach that is relatively efficient is to perform robust averaging of pairwise motions between scans~\cite{Govindu2014,Torsello11multi-viewregistration,Arrigoni2016}, or perform some variants of pose-graph optimization~\cite{RRIS}, to produce tightly registered scans. Yet another approach is to use the well-known IRLS method~\cite{IRLSHollandWelsch} to efficiently optimize robust cost functions~\cite{Aftab2015}. The recent fast global registration method of~\cite{FGR} made use of the duality between robust estimators and line processes~\cite{black1996unification} to develop a fast approach for global registration. This method has been shown to produce the best results till date in terms of speed as well as accuracy.

\section{Lie Group Structure of Euclidean Motions}
Our method utilizes the geometric structure of Lie groups, specifically that of the Special Euclidean group $\mathbb{SE}(3)$. In this section, we provide a brief description of the group structure and relevant Lie group properties. The Euclidean motion between two 3D scans consists of a rotation and a translation. While rotations can be represented in a variety of ways, we represent them as $3 \times 3$ orthonormal matrices $\bfR$, \textit{i.e.}, $\bfR \bfR^{\top}=\bfI_3$ and $|\bfR|=+1$ (here and for the rest of this paper, we use $\bfI_n$ to denote the $n \times n$ identity matrix). A rigid Euclidean motion $(\bfR,\bft)$ $(\bft \in \bbR^3)$ can then be compactly represented by a $4 \times 4$ matrix

\begin{equation} \label{Eqn:SE3MatrixForm}
\bfM=\left[\begin{array}{ccc}\bfR&|&\bft\\\hline\mathbf{0}&|&1\end{array}\right].
\end{equation}

\noi The matrices $\bfR$ and $\bfM$ satisfy the properties of a matrix group and also form smooth, differential manifolds, \textit{i.e.}, they are Lie groups. Thus, $\bfR \in \mathbb{SO}(3)$ and $\bfM \in \mathbb{SE}(3)$ where $\mathbb{SO}(3)$ and $\mathbb{SE}(3)$ are the Special Orthogonal and the Special Euclidean groups respectively. We also note that $\bfR$ and $\bfM$ have $3$ and $6$ degrees of freedom respectively.\\

\noi \textbf{Lie Groups: } The Lie group structure of $\mathbb{SE}(3)$ plays an important role in our formulation. Detailed expositions on the geometric properties of this representation are presented in ~\cite{Chirikjian02,SeligRobotics}. For our purposes, it will suffice to note that for finite dimensional Lie groups (matrix groups) the product and inverse operations are differentiable mappings. Every point in a Lie group has a local neighborhood (tangent space) called its Lie algebra which has the properties of a vector space. For $\bfR \in \mathbb{SO}(3)$ and $\bfM \in \mathbb{SE}(3)$, the corresponding Lie algebra are denoted as $[\boldsymbol{\omega}]_{\times} \in \mathfrak{so}(3)$ and $\mathfrak{m} \in \mathfrak{se}(3)$ respectively. Here $[\boldsymbol{\omega}]_{\times} \in \mathfrak{so}(3)$ is the skew-symmetric form of the axis-angle rotation representation $\boldsymbol{\omega}$. In this representation, $\bfR$ represents a rotation by angle $||\boldsymbol{\omega}||$ about the axis $\frac{\boldsymbol\omega}{||\boldsymbol{\omega}||}$. Further, we can move from a Lie group to its Lie algebra and vice-versa using the logarithm and the exponential mappings respectively. Thus, $\bfR = \exp ([\boldsymbol{\omega}]_{\times})$ and $[\boldsymbol{\omega}]_{\times} = \log (\bfR)$. Similarly, we have $\bfM = \exp (\mathfrak{m})$ and $\mathfrak{m} = \log (\bfM)$ with the forms

\begin{equation} \label{Eqn:SE3_se3}
\begin{split}
\mathfrak{m} =\log (\bfM) = \left[\begin{array}{ccc}[\boldsymbol{\omega}]_{\times}&|&\bfu\\\hline\mathbf{0}&|&0\end{array}\right];\\
\bfM = \exp (\mathfrak{m}) = \sum_{k=0}^{\infty} \frac{\mathfrak{m}^k}{k!} = \left[\begin{array}{ccc}\bfR&|&\bft\\\hline\mathbf{0}&|&1\end{array}\right].
\end{split}
\end{equation}

\noi Further, we note that the $\exp (\cdot)$ and $\log (\cdot)$ mappings for $\mathbb{SO}(3)$ and $\mathbb{SE}(3)$ have closed form expressions that can be efficiently implemented. Specifically,
\begin{equation}
    \bfR = \exp ([\boldsymbol{\omega}]_{\times}) = \bfI_3 + \sin\theta[\boldsymbol{\omega}]_{\times} + (1-\cos\theta)[\boldsymbol{\omega}]_{\times}^2
\end{equation}
\begin{equation}
    \bft = \bfP\bfu \; \text{ for } \; \bfP = \bfI_3 + \frac{(1-\cos\theta)}{\theta^2}[\boldsymbol{\omega}]_{\times} + \frac{(\theta-\sin\theta)}{\theta^3}[\boldsymbol{\omega}]_{\times}^2 \label{Eqn:tEqualsPu}
\end{equation}
\noi where $\theta = ||\boldsymbol{\omega}||$. We also note that the special properties of the group structure of $\mathbb{SE}(3)$ are discussed in~\cite{park1995distance}.

\section{Robust Motion Estimation}
Given $\mathcal{S} \geq 3$ pairs of point correspondences $\{(\bfp^s,\bfq^s)|1 \leq s \leq \mathcal{S} \}$ between a pair of scans, we can solve for the motion $\bfM$ required for aligning the pair of scans. In the case of global approaches, where the motion can potentially be large, such point correspondences are obtained by matching geometric features such as FPFH~\cite{FPFH} whereas in iterative schemes like ICP, the correspondence of points is obtained by finding the nearest neighbor match on the second scan for a point on the first one. Due to a variety of sources of error, the correspondence pairs are not in perfect alignment or could be grossly incorrect, \textit{i.e.}, $e^s = \left\Vert \bfp^s - \bfM \bfq^s \right\Vert \neq 0$, where $e^s$ denotes the norm of the discrepancy in registration for the $s$-th correspondence pair $(\bfp^s,\bfq^s)$. When we assume that the individual errors have a zero-mean, iid Gaussian distribution, the optimal estimate for $\bfM$ is obtained by a least squares minimization and has a closed form~\cite{Umeyama1991}. However, since correspondences could be highly incorrect in practice, instead of a least squares formulation, we pose motion estimation as the optimization of a robust cost function
\beq \label{Eqn:RobustCost}
\min_{\bfM \in \mathbb{SE}(3)} \mathcal{C}(\bfM) = \sum_{s=1}^{\mathcal{S}} \rho\big(\left\Vert \bfp^s - \bfM \bfq^s \right\Vert\big)  = \sum_{s=1}^{\mathcal{S}} \rho(e^s(\bfM))
\eeq

\noi where $\rho(\cdot)$ is a robust loss function and we also note that the individual error terms $e^s(\bfM)$ are a function of the motion $\bfM$. The use of robust estimators is well studied in statistics as an M-estimation problem and has been successfully used in a variety of vision problems~\cite{meer1991robust}. However, in addition to robustness, in Eqn.~\ref{Eqn:RobustCost} we require our solution to satisfy the geometric constraints for $\bfM \in \mathbb{SE}(3)$. These requirements of robust estimation under geometric constraints are satisfied by our solution.

\subsection{Proposed Solution for Pairwise Registration}
We propose to minimize the cost function $\mathcal{C}(\bfM)$ in Eqn.~\ref{Eqn:RobustCost} in an iterative fashion. Let the estimate for $\bfM$ at the $(k-1)$-th iteration be denoted $\bfM(k-1)$. In the $k$-th iteration, let us update the motion matrix by $\Delta \bfM(k)$, \textit{i.e.}, $\bfM(k) = \Delta\bfM(k)\bfM(k-1)$. Noting that here the matrix multiplication is not commutative, our formulation uses a left-invariant metric on $\mathbb{SE}(3)$~\cite{park1995distance,ZacurLeftInvariant}. Using a first-order approximation for the motion update matrix $\Delta \bfM(k)$, we have 
\begin{align}
\Delta \bfM(k) &\approx \bfI_4 + \Delta\mathfrak{m}(k) \label{Eqn:FirstOrderBCH} \\
\Rightarrow \mathcal{C}(\bfM(k)) &= \sum_{s=1}^{\mathcal{S}} \rho\big(\left\Vert \bfp^s - \bfM(k) \bfq^s \right\Vert\big)
\end{align}
\begin{equation}
= \sum_{s=1}^{\mathcal{S}} \rho\big(\left\Vert \bfp^s - (\bfI_4 + \Delta\mathfrak{m}(k))\bfM(k-1)\bfq^s \right\Vert\big).
\end{equation}
\noi Here we note that the Lie algebra matrix $\Delta\mathfrak{m}(k)$ encodes the $6$ parameters that we need to estimate for the update $\Delta\bfM(k)$. We can obtain the vector representation for these $6$ parameters using the `vee' operator, \textit{i.e.}, $\mathfrak{v} = \Delta\mathfrak{m}^{\vee} = \begin{bmatrix} \boldsymbol\omega & \bfu \end{bmatrix}^{\top}$. Since we use a first-order approximation in Eqn.~\ref{Eqn:FirstOrderBCH}, the cost $\mathcal{C}(\bfM(k))$ is linear in $\Delta\mathfrak{m}(k)$. We equivalently note that it is also linear in $\mathfrak{v}(k)$. Thus, we rewrite the individual error terms $e^s(\bfM(k))$ as

\begin{align}\label{Eqn:LinearModel}
\begin{split}
e^s(\bfM(k)) &= \left\Vert \bfp^s - (\bfI_4 + \Delta\mathfrak{m}(k))\bfM(k-1)\bfq^s \right\Vert \\
&= \left\Vert \bfA^s \mathfrak{v} - \bfb^s \right\Vert
\end{split}
\end{align}
\noi where $\bfA^s$ and $\bfb^s$ are the appropriate matrices. The derivation of the explicit forms of $\bfA^s$ and $\bfb^s$ are given in the appendix. To obtain the update in the $k$-th iteration, we now optimize the cost $\mathcal{C}(\mathbf{M}(k)) = \sum_{s=1}^{\mathcal{S}} \rho(e^s(\mathfrak{v}))$ w.r.t. $\mathfrak{v}$, and get,

\begin{align}
\frac{\rho^{\prime}(e^s)}{e^s} (\bfA^s)^{\top} \bfA^s \mathfrak{v} &= \frac{\rho^{\prime}(e^s)}{e^s} (\bfA^s)^{\top} \bfb^s \label{Eqn:Weights}
\end{align}

\noi for each summand indexed by $s$, where $\rho^{\prime}(\cdot) = \frac{\partial \rho}{\partial e}$ is the influence function of the robust loss $\rho(\cdot)$. We may further denote $w^s = \frac{\rho^{\prime}(e^s)}{e^s}$, which is the relative weight accorded to the $s$-th equation in Eqn.~\ref{Eqn:Weights}. Collecting all such relationships obtained for each pair of correspondences $(\bfp^s,\bfq^s)$ into a single system of equations we have

\begin{equation}\label{Eqn:LinearSystem}
    \bfA^{\top} \bfW \bfA \mathfrak{v} = \bfA^{\top} \bfW \bfb
\end{equation}
where $\bfA = \left[\begin{array}{c}\bfA^1 \\ \dots \\ \bfA^{\mathcal{S}} \end{array} \right]$, $\bfb =\left[\begin{array}{c} \bfb^1\\ \dots \\ \bfb^{\mathcal{S}}\end{array} \right]$, and $\bfW = \left[\begin{array}{ccc}w^1 \bfI_3 & &\\&\ddots & \\ & & w^{\mathcal{S}} \bfI_3 \end{array} \right]$.\\
\indent Eqn.~\ref{Eqn:LinearSystem} is a weighted linear system of equations with the solution $\mathfrak{v} = {(\bfA^{\top} \bfW \bfA)}^{-1} \bfA^{\top} \bfW \bfb$. However, it should be noted that each individual weight $w^s$ is a function of the error $e^s$ which, in turn, is dependent on $\mathfrak{v}$, since $e^s = \left\Vert \bfA^s \mathfrak{v} - \bfb^s \right\Vert$. Thus, in Eqn.~\ref{Eqn:LinearSystem}, the equivalent relationship is $\bfA^{\top} \bfW(\mathfrak{v}) \bfA \mathfrak{v} = \bfA^{\top} \bfW(\mathfrak{v}) \bfb$. The solution for this system of equations is the well-known iteratively reweighted least squares (IRLS) method~\cite{IRLSHollandWelsch,SzeliskiBook}. In the IRLS method, in each iteration the weights $w^s$ are estimated based on the current estimate of $\mathfrak{v}$. Given these weights, $\mathfrak{v}$ is re-estimated. This process is repeated till convergence.\\
\indent Given a solution for $\mathfrak{v}$, we can estimate $\Delta\bfM(k) = \exp (\hat{\mathfrak{v}}(k))$ where the `hat' operator $\hat{\mathfrak{v}}$ converts the estimated Lie algebra parameters $\mathfrak{v}$ into its equivalent matrix representation $\Delta\mathfrak{m}(k)$. We emphasize here that although in Eqn.~\ref{Eqn:FirstOrderBCH}, we assumed a first-order approximation, we map the estimated $\mathfrak{v}(k)$ into an intrinsic estimate of the motion update, \textit{i.e.}, $\Delta\bfM(k) = \exp (\hat{\mathfrak{v}}(k)) \in \mathbb{SE}(3)$. In other words, a first-order approximation in an intermediate step does not mean that the actual cost function is approximated. The mapping $\Delta\mathfrak{m}(k) = \exp(\hat{\mathfrak{v}}(k))$ ensures that the estimated motion $\bfM(k)$ is always a valid member of the $\mathbb{SE}(3)$ group. We may now state our solution for robust motion estimation as given in Algorithm~\ref{Algo:PairwiseIRLS}.\\

\begin{algorithm}[h]
\caption{IRLS estimation of pairwise 3D registration}\label{Algo:PairwiseIRLS}
Input: $\{(\bfp^1,\bfq^1) \cdots (\bfp^{\mathcal{S}}, \bfq^{\mathcal{S}})\}$ ($\mathcal{S}$ correspondences across a pairs of scans) \\
Output: $\bfM \in \mathbb{SE}(3)$  (Robust estimate of motion between scans) \\
Initialization: $\bfM$ is set to $4 \times 4$ identity matrix

\begin{algorithmic}[l]
\While {$||\mathfrak{v}|| > \epsilon$}
\State 1. Compute $\{(\bfA^s,\bfb^s)| \forall s \in [1 \cdots {\mathcal{S}}]\}$ using Eqn.~\ref{Eqn:LinearModel}
\State 2. Compute weights $w^s = \frac{\rho^{\prime}(e^s)}{e^s}$ as defined by Eqn.~\ref{Eqn:Weights}
\State 3. Estimate $\mathfrak{v}$ as IRLS solution for Eqn.~\ref{Eqn:LinearSystem} 
\State 4. Update $\bfM \leftarrow \exp(\hat{\mathfrak{v}}) \bfM$
\EndWhile
\end{algorithmic}
\end{algorithm}
\indent Algorithm~\ref{Algo:PairwiseIRLS} is an iterative algorithm with a nested iteration. The outer loop is defined by the \texttt{while} statement and we denote its number of iterations as $K_{outer}$. The inner loop consists of the IRLS step of line 3 in Algorithm~\ref{Algo:PairwiseIRLS} since IRLS is itself an iterative method. We denote the number of iterations of the IRLS step as $K_{IRLS}$.

\subsection{Extension to Joint Multiview Registration}\label{Sec:Multiview}
In Algorithm~\ref{Algo:PairwiseIRLS} we presented the registration solution for two scans. This approach can be extended to the simultaneous or joint multiview registration of a set of scans. Towards this end, we define a viewgraph $\mathcal{G} = \{\mathcal{V},\mathcal{E}\}$ where $v_i \in \mathcal{V}$ represents the Euclidean motion of the $i$-th scan (equivalently camera) and an edge $(i,j) \in \mathcal{E}$ signifies that the relative Euclidean motion between the $i$-th and $j$-th scan can be determined from available matches. Further, we denote the number of scans as $N = |\mathcal{V}|$. We may now define the cost function to be optimized for joint multiview registration as follows
\begin{align} \label{Eqn:RobustCostMV}
\begin{split}
\mathcal{C}(\bbM) &= \sum_{(i, j) \in \mathcal{E}}\sum_{s=1}^{\mathcal{S}_{ij}} \rho\big(\left\Vert \bfM_i\bfp_i^s - \bfM_j\bfp_j^s \right\Vert\big)\\
&= \sum_{(i, j) \in \mathcal{E}}\sum_{s=1}^{\mathcal{S}_{ij}} \rho(e_{ij}^s(\bbM))
\end{split}
\end{align}
\noi where $\bbM = \left\{ \bfM_1 \cdots \bfM_N \right\}$ denotes the set of absolute motions of each of the $N$ scans w.r.t. to a global frame of reference and $\mathcal{S}_{ij}$ is the number of correspondences between the $i$-th and $j$-th scan. We again use an iterative approach to minimize the cost function $\mathcal{C}(\bbM)$ in Eqn.~\ref{Eqn:RobustCostMV} w.r.t. $\bbM$. In the $k$-th iteration, we update each motion matrix $\bfM_i$ by $\Delta\bfM_i(k)$, \textit{i.e.}, $\bfM_i(k) = \Delta\bfM_i(k)\bfM_i(k-1)$ $\forall i \in [1 \cdots N]$. Using a first-order approximation for each update matrix $\Delta\bfM_i(k)$, we have
\begin{align}\label{Eqn:LinearModelMV}
\begin{split}
e_{ij}^s(\bbM(k)) &= \lVert (\bfI_4 + \Delta\mathfrak{m}_i(k))\bfM_i(k-1)\bfp_i^s \\
&\quad - (\bfI_4 + \Delta\mathfrak{m}_j(k))\bfM_j(k-1)\bfp_j^s \rVert
\end{split} \\
&= \left\Vert \bfA_{ij}^s \mathbbm{v} - \bfb_{ij}^s \right\Vert
\end{align}
\noi where $\mathbbm{v} = \begin{bmatrix} \mathfrak{v}_1 & \cdots & \mathfrak{v}_N \end{bmatrix}^{\top}$ collates the corresponding vector representations of each of the Lie algebra matrices $\Delta\mathfrak{m}_i(k)$, and $\bfA_{ij}^s$ and $\bfb_{ij}^s$ are constructed analogous to Eqn.~\ref{Eqn:LinearModel}. The subsequent update in the $k$-th iteration is then analogously obtained from the relation
\begin{equation}\label{Eqn:LinearSystemMV}
\bbA^{\top} \bbW(\mathbbm{v}) \bbA \mathbbm{v} = \bbA^{\top} \bbW(\mathbbm{v}) \mathbbm{b}
\end{equation}
where $\bbA$, $\mathbbm{b}$ and $\bbW(\mathbbm{v})$ correspondingly collate all $\bfA_{ij}^s$, $\bfb_{ij}^s$ and $\bfw_{ij}^s = \frac{\rho^{\prime}(e_{ij}^s)}{e_{ij}^s}$. As earlier, the solution for the system of equations in Eqn.~\ref{Eqn:LinearSystemMV} is the IRLS method, where we estimate the weights $w_{ij}^s$ and the collated vector representation $\mathbbm{v}$ in alternating iterations till convergence.\\
\indent Given a solution for $\mathbbm{v}$, we can estimate, for each $i \in \left[ 1 \cdots N \right]$, $\Delta\bfM_i(k) = \exp (\hat{\mathfrak{v}_i}(k))$ and thereby update each member of $\bbM(k)$ as $\bfM_{i}(k) \leftarrow \Delta\bfM_{i}(k)\bfM_{i}(k-1)$. It should also be noted that in multiview registration, the choice of the global frame of reference for the set $\bbM$ is arbitrary. For our implementation, we fix it to the first scan without loss of generality, \textit{i.e.}, we set $\bfM_1 = \bfI_4$ and do not update $\bfM_1$ throughout course of optimization. The detailed algorithm is presented in the appendix.

\section{Results}\label{Sec:Results}
In recent literature, the fast global registration (FGR) method of ~\cite{FGR} has been shown to outperform other global registration and local refinement algorithms in terms of speed and registration accuracy. Therefore, owing to space constraints, we confine the comparison of our results mostly to those of FGR as this will suffice to show where our performance stands in terms of the other registration algorithms. Furthermore, we test our algorithm on 3 different choices of the loss function $\rho(\cdot)$, namely $L_{\frac{1}{2}}$: $\rho(x) = \sqrt{\lvert x \rvert}$, $L_{1}$: $\rho(x) = \lvert x \rvert$, and scaled Geman-McClure: $\rho(x) = \frac{\mu x^2}{\mu + x^2}$, where $\mu$ is the scale factor. The FGR approach of~\cite{FGR} uses only the scaled Geman-McClure loss function and anneals $\mu$ in fixed steps per iteration. To enable comparison, in our tests with the scaled Geman-McClure loss function we anneal $\mu$ in an identical manner. We present results on both pairwise and multiview registration tests. For these pairwise and multiview tests we terminate using $\epsilon = {10}^{-5}$ and $\epsilon={10}^{-7}$ respectively. All our reported running times are measured on a single thread on an Intel Xeon(R) E5-2650 v3 processor clocked at 2.30 GHz.

\subsection{Pairwise Registration}
We present the performance of our proposed pairwise registration on the synthetic range data provided by~\cite{FGR} and on the 4 indoor sequences in the Augmented ICL-NUIM dataset provided by Choi et al.~\cite{RRIS}. In all our pairwise tests, we use $K_{IRLS} = 2$. We compare the registration errors given by the 3 versions of our method with the following prior methods: Super4PCS~\cite{Mellado2014}, GoICP~\cite{Yang2016}, Choi et al.~\cite{RRIS}, FGR~\cite{FGR} and DARE~\cite{DARE} (using the hyperparameters suggested by the authors). Registration errors comprise of statistical measures on the distances between the ground-truth point correspondences between pairs of scans post alignment. We also report the trajectory errors, which include statistical measures on both the rotation errors and the translation norm errors between the pairs of cameras (corresponding to the given scans) w.r.t. the ground-truth camera pair, for all the methods.\\
\noi \textbf{Synthetic range dataset:} We perform the set of controlled experiments described in~\cite{FGR} on each of their 5 synthetic datasets, at the given Gaussian noise levels $\sigma = 0.0025$ and $\sigma = 0.005$ (for each model, $\sigma$ is in units of the surface diameter). Note that adding Gaussian noise to a pair of scans introduces outliers in the point correspondences that are computed between that pair. Since depth cameras produce noisy scans in practice, this is a realistic way of increasing the outlier percentage in the point correspondences for synthetic scans. Table~\ref{Table:RegSynthetic} lists for each method and for each noise level, the mean and maximal RMSE on the aligned ground truth correspondences. For both noise levels, our method with the $L_{\frac{1}{2}}$ loss function attains the lowest registration error together with FGR. Table~\ref{Table:RunningTimeSynthetic} reports the mean running time of the motion step of each method for each of the 5 models in the dataset. The $L_{\frac{1}{2}}$ of our method is more $3\times$ faster than all prior methods on the average, and the $L_1$ version is more than $5\times$ faster.

\begin{table}[t]
    \centering
    \resizebox{\linewidth}{!}{%
    \begin{tabular}{| l || c | c | c | c || c | c | c | c |}
        \hline
        \multirow{2}{*}{Method} & \multicolumn{4}{c||}{$\sigma = 0.0025$} & \multicolumn{4}{c|}{$\sigma = 0.0050$} \\
        \cline{2-9}
        & Md.RAE & Md.TNE & Mn.RMSE & Mx.RMSE & Md.RAE & Md.TNE & Mn.RMSE & Mx.RMSE \\
        \hline
        Super4PCS~\cite{Mellado2014} & 0.864 & 0.008 & 0.014 & 0.029 & 1.468 & 0.012 & 0.017 & 0.095 \\
        \hline
        GoICP~\cite{Yang2016} & 1.207 & 0.011 & 0.032 & 0.133 & 1.736 & 0.019 & 0.037 & 0.127 \\
        \hline
        Choi et al.~\cite{RRIS} & 0.778 & 0.006 & 0.008 & 0.022 & 1.533 & 0.015 & 0.035 & 0.274 \\
        \hline
        FGR~\cite{FGR} & 0.749 & 0.005 & 0.004 & 0.011 & 1.146 & 0.008 & 0.006 & 0.017 \\
        \hline
        DARE~\cite{DARE} & 1.851 & 0.013 & 0.035 & 0.176 & 2.005 & 0.025 & 0.054 & 0.312 \\
        \hline
        Our $L_{\frac{1}{2}}$ & \textbf{0.545} & \textbf{0.004} & \textbf{0.004} & \textbf{0.011} & \textbf{0.959} & \textbf{0.008} & \textbf{0.006} & \textbf{0.017} \\
        \hline
        Our $L_{1}$ & 0.566 & 0.004 & 0.004 & 0.011 & 1.516 & 0.011 & 0.007 & 0.017 \\
        \hline
        Our GM & 0.725 & 0.005 & 0.004 & 0.011 & 1.146 & 0.008 & 0.006 & 0.017 \\
        \hline
    \end{tabular}%
    }
    \caption{Median rotation angle error (Md.RAE) (in degrees), median translation norm error (Md.TNE), mean RMSE (Mn.RMSE) and maximal RMSE (Mx.RMSE) (all in units of the surface diameters) achieved by each method for each noise level $\sigma$ on the synthetic range datasets.}
    \label{Table:RegSynthetic}
\end{table}

\begin{table}[t]
    \centering
    \resizebox{\linewidth}{!}{%
    \begin{tabular}{| l | c | c | c | c | c | c | c | c | c | }
        \hline
        Dataset & Mean \# points & Super 4PCS \cite{Mellado2014} & GoICP \cite{Yang2016} & Choi et al. \cite{RRIS} & FGR \cite{FGR} & DARE \cite{DARE} & Our $L_{\frac{1}{2}}$ & Our \newline $L_{1}$ & Our GM \\
        \hline
        Bimba & 9,416 & 16,230 & 1,550 & 650 & 11.9 & 920 & 3.9 & \textbf{2.5} & 5.5 \\
        \hline
        Child'n & 11,148 & 18,410 & 1,620 & 890 & 16.8 & 960 & 4.8 & \textbf{3.3} & 8.0 \\
        \hline
        Dragon & 11,232 & 20,520 & 1,840 & 970 & 17.6 & 1,090 & 5.0 & \textbf{3.4} & 8.2 \\
        \hline
        Angel & 12,072 & 29,640 & 3,000 & 1,090 & 19.1 & 1,770 & 5.3 & \textbf{3.8} & 8.9 \\
        \hline
        Bunny & 13,357 & 38,470 & 5,530 & 1,170 & 21.6 & 3,310 & 7.4 & \textbf{5.1} & 9.9 \\
        \hline Mean & 11,445 & 24,650 & 2,710 & 960 & 17.4 & 1,610 & 5.3 & \textbf{3.6} & 8.1 \\
        \hline
    \end{tabular}%
    }
    \caption{Running time (in milliseconds) of the motion step of each method for each model in the synthetic range dataset.}
    \label{Table:RunningTimeSynthetic}
\end{table}
\noi \textbf{Augmented ICL-NUIM dataset:} Each of the 4 sequences in the Augmented ICL-NUIM dataset provided by Choi et al.~\cite{RRIS} consist of 2350 to 2870 scans of an indoor scene. Moreover, the given scans are provided in a smooth temporal sequence, \textit{i.e.}, pairs of scans with proximity in timestamp also have sufficient view overlap with each other. This, in turn, leads to reliable FPFH feature matches between such pairs. We therefore tested the performance of all the methods for all pairs of scans $(i, j)$ in each sequence such that $\lvert i-j \rvert \leq 10$. Table~\ref{Table:Traj&TimeICL} lists the results on each dataset for the various methods under consideration. For each dataset and corresponding method, we list the median rotation angle error and the median translation norm error of the recovered pairwise camera motions, as well as the mean computation time on the pairs. The $L_{\frac{1}{2}}$ version of our method performs the best in terms of the trajectory error statistics. It can also be seen to be significantly faster than the FGR method of~\cite{FGR}. More such pairwise results on other datasets are given in the appendix.

\begin{table}[t]
    \centering
    \resizebox{\linewidth}{!}{%
    \begin{tabular}{| l || c|c|c || c|c|c || c|c|c || c|c|c | }
        \hline
        \multirow{2}{*}{Method} & \multicolumn{3}{c||}{\texttt{livingroom 1}} & \multicolumn{3}{c|}{\texttt{livingroom 2}} & \multicolumn{3}{c||}{\texttt{office 1}} & \multicolumn{3}{c|}{\texttt{office 2}} \\
        \cline{2-13}
         & Md.RAE & Md.TNE & Mn.Time & Md.RAE & Md.TNE & Md.Time & Md.RAE & Md.TNE & Mn.Time & Md.RAE & Md.TNE & Mn.Time \\
        \hline
        Super4PCS~\cite{Mellado2014} & 1.104 & 0.039 & 368,030 & 0.616 & 0.033 & 344,720 & 0.932 & 0.038 & 367,980 & 0.844 & 0.027 & 345,460 \\
        \hline
        GoICP~\cite{Yang2016} & 1.336 & 0.071 & 35,110 & 0.992 & 0.058 & 33,420 & 1.365 & 0.066 & 34,450 & 1.104 & 0.047 & 32,530  \\
        \hline
        Choi et al.~\cite{RRIS} & 0.941 & 0.041 & 14,740 & 0.551 & 0.031 & 13,850 & 0.811 & 0.036 & 14,720 & 0.765 & 0.029 & 13,990 \\
        \hline
        FGR~\cite{FGR} & 0.793 & 0.029 & 272 & 0.482 & 0.021 & 181 & 0.707 & 0.020 & 272 & 0.669 & 0.016 & 177 \\
        \hline
        DARE~\cite{DARE} & 1.305 & 0.044 & 21,500 & 1.172 & 0.059 & 20,320 & 1.716 & 0.037 & 21,110 & 1.286 & 0.068 & 20,920 \\
        \hline
        Our $L_{\frac{1}{2}}$ & \textbf{0.595} & \textbf{0.023} & 61 & \textbf{0.380} & \textbf{0.017} & 50 & \textbf{0.474} & \textbf{0.014} & 59 & \textbf{0.437} & \textbf{0.011} & 45 \\
        \hline
        Our $L_{1}$ & 0.964 & 0.025 & \textbf{33} & 0.419 & 0.019 & \textbf{27} & 0.569 & 0.017 & \textbf{33} & 0.524 & 0.013 & \textbf{25} \\
        \hline
        Our GM & 0.793 & 0.029 & 118 & 0.482 & 0.021 & 87 & 0.707 & 0.020 & 118 & 0.669 & 0.016 & 89 \\
        \hline
    \end{tabular}%
    }
    \caption{Median rotation angle error (RAE) (in degrees), median translation norm error (TNE) (in meters) and mean running time (in milliseconds) of the motion step of each method for each sequence in the Augmented ICL-NUIM dataset.}
    \label{Table:Traj&TimeICL}
\end{table}

\subsection{Joint Multiview Registration}
\begin{figure}[t]
    \centering
    \includegraphics[width=\linewidth]{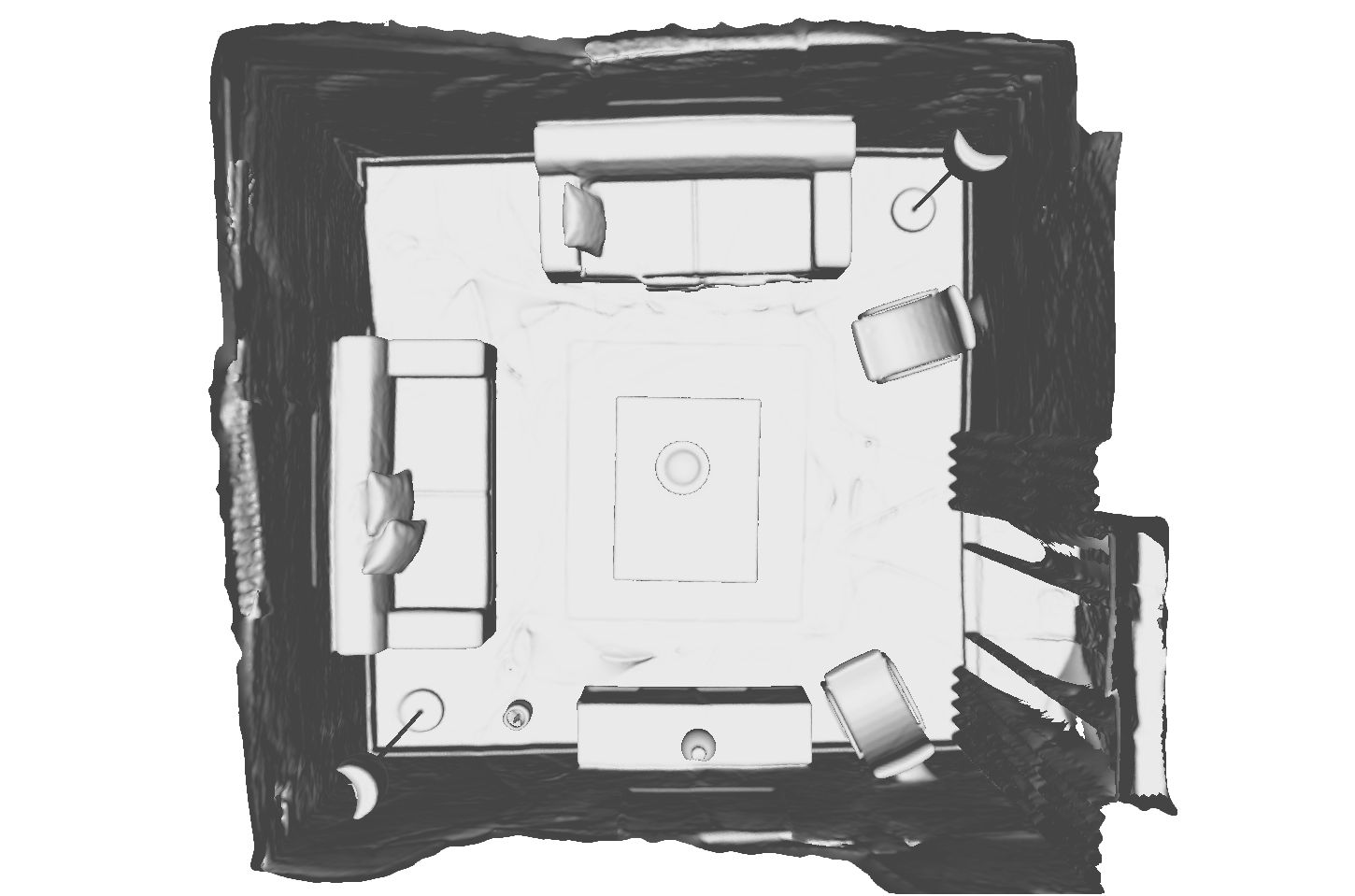}
    \caption{Reconstruction of the \texttt{livingroom 2} sequence from the Augmented ICL-NUIM dataset, as given by our method with the $L_{\frac{1}{2}}$ loss function}
    \label{Fig:Lvr2nRecons}
\end{figure}

We present the performance of our joint multiview registration algorithm on the 4 sequences in the Augmented ICL-NUIM dataset, specifically, on the 47 to 57 scene fragments that were provided for each sequence by Choi et al.~\cite{RRIS}. We use $K_{IRLS} = 3$ for multiview registration, as the joint optimization variable and its corresponding search space are both large ($\mathbbm{v} \in \mathbb{R}^{6(N-1)}$ in Eqn.~\ref{Eqn:LinearSystemMV} for $N$ cameras). First, we compute pairwise motion estimates between fragments followed by a robust motion averaging step on $\mathbb{SE}(3)$ that is similar to that used for rotation averaging in~\cite{RRAPAMI}. The output of this two-stage approach is used to initialize the joint multiview optimization in Eqn.~\ref{Eqn:RobustCostMV}. The main drawback of only using the two-step approach for global registration is that it is not a true global method. It only considers local point correspondences, and then averages out the errors made by the pairwise motion estimates. Conversely, the joint multiview approach deals with point correspondences in the global setting and solves for the global cost function. The relative improvement in reconstruction error gained by using the joint multiview approach on top of the two-stage approach is shown in a table in the appendix.\\
\indent We compare our results with those of Choi et al.~\cite{RRIS} and FGR~\cite{FGR}. While we are aware of other approaches to multiview registration including closed form approaches, we omit them from our comparisons because they failed to provide a solution in the large-scale dataset for multiview registration we have used. For example, Bartoli et al.~\cite{Bartoli2013} and Bergstr{\"o}m et al.~\cite{Bergstrom2014}, among others, compute transformations using a closed form SVD solution that do not scale to large scale data. Other alternates, such as that of Fitzgibbon et al.~\cite{Fitzgibbon2003}, use an LM-based approach, which are slow and do not exploit the geometry of the problem.\\
\indent Table~\ref{Table:Traj&RegMV} lists the mean registration error from the ground truth surface achieved by each compared method on each sequence as well as the time taken to complete execution. For estimating the mean registration error, we use the CloudCompare software available at \url{http://www.cloudcompare.org}.\\
\indent Once again, the $L_{\frac{1}{2}}$ version of our method performs the best overall in terms of registration error and is significantly faster than the FGR method of~\cite{FGR}. Also our $L_{1}$ method is the fastest amongst all methods with a slight drop in accuracy compared to our $L_{\frac{1}{2}}$ method. Finally, Figure~\ref{Fig:Lvr2nRecons} shows a complete reconstruction of the sequence \texttt{livingroom 2} from the Augmented ICL-NUIM dataset produced by the $L_{\frac{1}{2}}$ version of our multiview method. Reconstructions of other sequences are given in the appendix.\\

\begin{table}[t]
    \centering
    \begin{tabular}{| l || c | c || c | c || c | c || c | c | }
        \hline
        \multirow{2}{*}{Method} & \multicolumn{2}{c||}{\texttt{livingroom 1}} & \multicolumn{2}{c||}{\texttt{livingroom 2}} & \multicolumn{2}{c||}{\texttt{office 1}} & \multicolumn{2}{c|}{\texttt{office 2}} \\
        \cline{2-9}
        & MRE & Time & MRE & Time & MRE & Time & MRE & Time \\
        \hline
        Choi et al.~\cite{RRIS} & 0.04 & 8,940 & 0.07 & 3,360 & 0.03 & 4,500 & 0.04 & 4,080 \\
        \hline
        FGR~\cite{FGR} & 0.05 & 131 & 0.06 & 81 & 0.03 & 69 & 0.05 & 48 \\
        \hline
        Our $L_{\frac{1}{2}}$ & \textbf{0.04} & 71 & \textbf{0.05} & 49 & \textbf{0.03} & 42 & \textbf{0.04} & 32 \\
        \hline
        Our $L_{1}$ & 0.07 & \textbf{62} & 0.09 & \textbf{40} & 0.04 & \textbf{36} & 0.06 & \textbf{28} \\
        \hline
        Our GM & 0.05 & 88 & 0.06 & 70 & 0.03 & 55 & 0.05 & 41 \\
        \hline
    \end{tabular}
    \caption{Mean registration error (MRE) (in meters) and running time (in seconds) for each method for full reconstruction from the fragments of each sequence in the Augmented ICL-NUIM dataset.}
    \label{Table:Traj&RegMV}
\end{table}
\vspace{-20pt}
\section{Discussion}
As we demonstrated in Section~\ref{Sec:Results}, our motion estimation method in Algorithm~\ref{Algo:PairwiseIRLS} is both fast and accurate. More specifically, our method outperforms the state-of-the-art FGR method of~\cite{FGR} in terms of both speed and accuracy. Given their strong similarities, in Section~\ref{subsec:FGR_compare} we examine the relationship of the method of~\cite{FGR} and our approach in Algorithm~\ref{Algo:PairwiseIRLS}. Following that, we discuss the limitations of using FPFH for feature matching and our approach to overcome those in Section~\ref{subsec:FPFH_limitations}.
\subsection{Comparison with FGR~\cite{FGR}}\label{subsec:FGR_compare}
\begin{figure}
\centering
\begin{minipage}{0.5\textwidth}
    \centering
    \includegraphics[width=\linewidth]{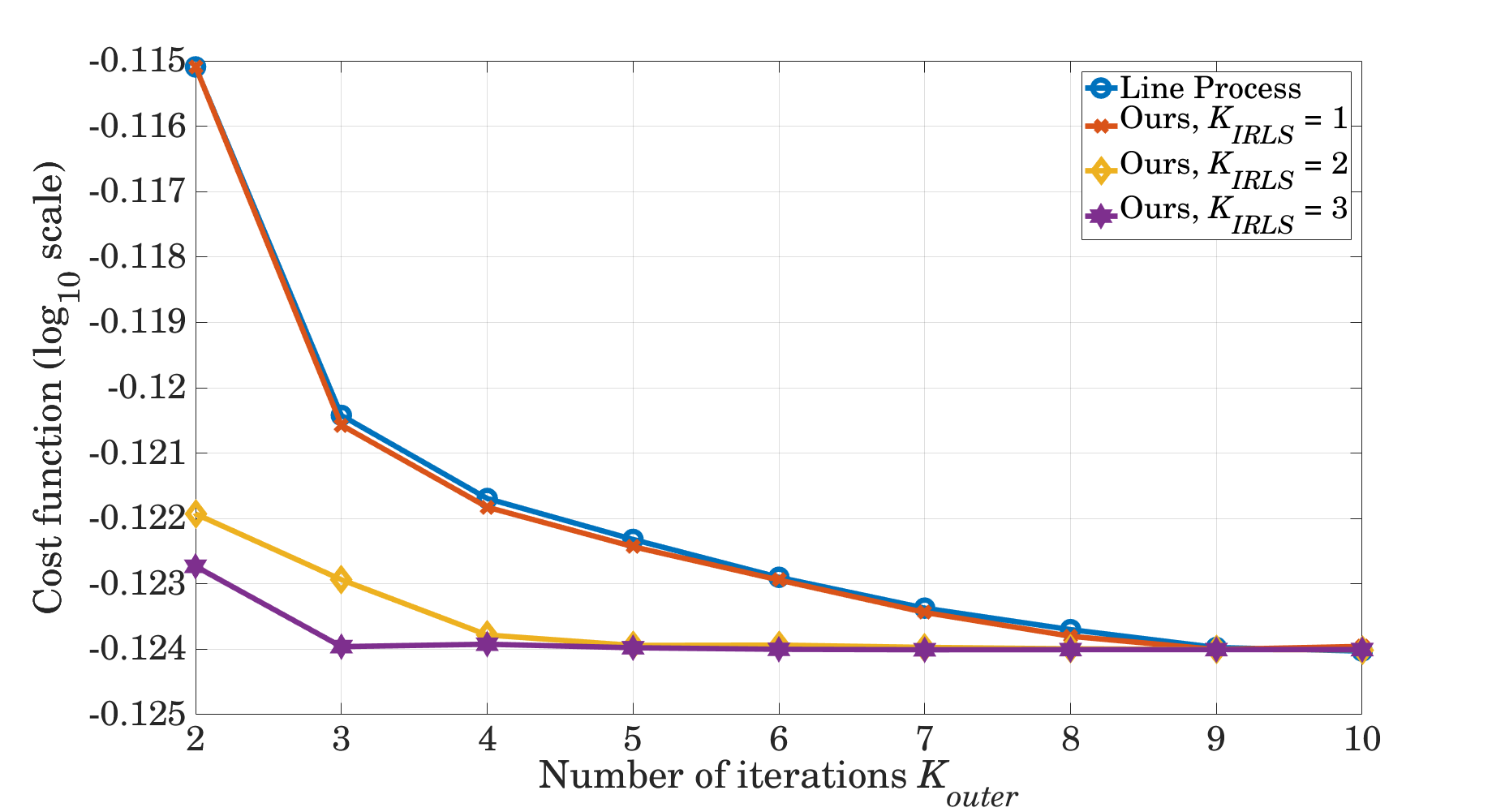}
    \caption{Comparison of the line process solution with our method for different $K_{IRLS}$. For ease of visualization, we show performance only for the iterations between $2$ and $10$.}
	\label{Fig:ConvGraph}
\end{minipage}
\hfill
\begin{minipage}{0.48\textwidth}
    \centering
    \captionsetup{type=table} 
	\begin{tabular}{|c|C{0.6cm}|C{0.6cm}|C{0.6cm}|C{0.6cm}|}
	\hline
	\multirow{2}{*}{$\epsilon$} & LP & \multicolumn{3}{c|}{Ours, $K_{IRLS}=$} \\
    	\cline{3-5}
    	&~\cite{FGR} & $1$ & $2$ & $3$ \\
    	\hhline{|=|=|=|=|=|}
    	$10^{-1}$ & $2$ & $2$ & $2$ & $2$ \\
    	\hline
    	$10^{-2}$ & $5$ & $5$ & $4$ & $4$ \\
    	\hline
    	$10^{-3}$ & $13$ & $13$ & $8$ & $6$ \\
    	\hline
    	$10^{-4}$ & $21$ & $18$ & $11$ & $8$ \\
    	\hline
    	$10^{-5}$ & $27$ & $24$ & $13$ & $10$ \\
    	\hline
    	$10^{-6}$ & $33$ & $32$ & $19$ & $14$ \\
    	\hline
    	$10^{-7}$ & $63$ & $46$ & $25$ & $18$ \\
    	\hline
	\end{tabular}
	\caption{Number of iterations $K_{outer}$ taken by each method to reach each convergence criterion $\epsilon$}
	\label{Table:ConvList}
\end{minipage}
\end{figure}

In~\cite{FGR}, the cost function to be minimized is the same as that of Eqn.~\ref{Eqn:RobustCost}. However, for minimizing this cost function, they use a line process optimization. Originally developed in~\cite{black1996unification} for modelling discontinuities, a line process optimization can be shown to be equivalent to optimizing a robust estimator. Recalling that $e^s = \left\Vert \bfp^s - \bfM \bfq^s \right\Vert$, we define a cost function
\begin{equation}\label{Eqn:LineProcess}
E(\bfM,\bfL) = \sum_{s=1}^{\mathcal{S}} (e^s)^{2} l^s + \psi(l^s) 
\end{equation}
\noi where $\bfL = [l^1 \cdots l^{\mathcal{S}}]$ is the collection of line processes $l^s$ for each correspondence pair $(\bfp^s,\bfq^s)$ and $\psi(\cdot)$ is a penalty term for each line process. Here $\psi(l)$ is a monotonically decreasing function designed such that when $l=0$, $\psi(l)$ is a fixed non-negative constant and when $l=1$, $\psi(l)=0$. Thus, varying the line process $l$ in the interval $[0,1]$ allows us to move between a least-squares and a robust regime. In~\cite{black1996unification} it has been shown that for every choice of loss function $\rho(\cdot)$, there is an equivalent $\psi(\cdot)$ such that minimization in Eqn.~\ref{Eqn:LineProcess} yields the same solution as that of Eqn.~\ref{Eqn:RobustCost}. The FGR method of~\cite{FGR} utilizes the robustness of the line process to estimate the desired $\bfM \in \mathbb{SE}(3)$. Using a first-order approximation of the motion update,~\cite{FGR} arrives at a Gauss-Newton solution (Eqn. 8 of~\cite{FGR}). It can be easily shown that this solution is identical to solving the system of equations in Eqn.~\ref{Eqn:LinearSystem} in our notation. In other words, while solving for the update step $\mathfrak{v}(k)$, the FGR method of~\cite{FGR} implicitly carries out a \textbf{single} iteration of our IRLS step in line 3 of Algorithm~\ref{Algo:PairwiseIRLS}. In contrast, in our approach we carry out $K_{IRLS}>1$ iterations of the IRLS step to achieve better convergence.\\
\indent Although the difference between our method and that of~\cite{FGR} is only in the number of iterations of the inner IRLS step, its implication is both subtle and profound. If we were solving a single IRLS problem in a vector space setting, this difference would have been immaterial. However, we note that in both of our approaches, the linear solution for the updates $\mathfrak{v}(k)$ are interleaved with the non-linear motion updates $\bfM \leftarrow \exp(\hat{\mathfrak{v}}) \bfM$, resulting in significantly different trajectories of the solution $\bfM(k)$ on the $\mathbb{SE}(3)$ group. Specifically, in our case, by iterating the IRLS step to convergence we obtain the best possible estimate of $\mathfrak{v}(k)$ in each intermediate step which, in turn, results in the best improvement of the estimate $\bfM(k)$ (equivalently the most reduction in the cost function $\mathcal{C}(\cdot)$ in the $k$-th step).\\
\indent Another noteworthy difference between the two methods is the choice of the parametrization of the motion representation. We use the geometrically correct form of $\Delta\mathfrak{m}(k)$ in Eqn.~\ref{Eqn:FirstOrderBCH}, \textit{i.e.}, $\mathfrak{v} = \begin{bmatrix} \boldsymbol\omega & \bfu \end{bmatrix}^{\top}$ for our update step. However, for their update step, the authors of~\cite{FGR} use an extrinsic form of motion parametrization, \textit{i.e.}, $\begin{bmatrix} \boldsymbol\omega & \bft \end{bmatrix}^{\top}$. While our parametrization is geometrically consistent with Lie group theory, we can recognize from Eqn.~\ref{Eqn:tEqualsPu} that the choice in~\cite{FGR} is approximately close to our representation for small motion, \textit{i.e.}, $\bfu \rightarrow \bft$ if and only if $\theta \rightarrow 0$. Conversely, for sufficiently large $\theta$, the approximate representation $\begin{bmatrix} \boldsymbol\omega & \bft \end{bmatrix}^{\top}$ is notably different from the exact representation $\begin{bmatrix} \boldsymbol\omega & \bfu \end{bmatrix}^{\top}$ of the $\mathfrak{se}(3)$ form. Therefore the improvement per iteration in the cost function for the method of~\cite{FGR} is lower compared to our single iteration IRLS form. The result is that the line process-based method of~\cite{FGR} has slower convergence.\\
\indent We highlight both these differences for a pair of scans from the Augmented ICL-NUIM dataset in Figure~\ref{Fig:ConvGraph}. For the purpose of illustration, we consider the $L_{\frac{1}{2}}$ loss function for the line process optimization routine proposed in~\cite{FGR} as well as for our optimization routine. The reason we do not consider the Geman-McClure loss function is that it has a scale factor, which, in practice, is initialized at a large value and has to be progressively annealed during optimization. This annealing introduces artifacts that unnecessarily clutter the illustration of the underlying differences. In other words, we use $L_{\frac{1}{2}}$ because it does not alter the fundamental properties of the two optimization routines, and at the same time lends itself to a clean illustration of our argument. We also note that in Figure~\ref{Fig:ConvGraph}, we represent the cost $\mathcal{C}(\mathbf{M})$ as a function of the number of iterations $K_{outer}$, and that it is depicted on a $\log_{10}$ scale. For ease of visualization, we only show the plot for the iteration range $[2,10]$. Alongside the figure, Table~\ref{Table:ConvList} reports the number of iterations $K_{outer}$ taken by each method to reach the different convergence criteria specified by $\epsilon$.\\

\indent Firstly we observe from Table~\ref{Table:ConvList} that even though the line process is conceptually equivalent to our procedure with a single IRLS step, the proposed optimization routine of~\cite{FGR} takes more iterations than our actual procedure with a single IRLS step to converge to the same convergence criterion. This is because of the extrinsic (approximate) parametrization $\begin{bmatrix} \boldsymbol\omega & \bft \end{bmatrix}^{\top}$ used in~\cite{FGR} as opposed to the correct Lie algebraic representation $\begin{bmatrix} \boldsymbol\omega & \bfu \end{bmatrix}^{\top}$. Secondly, it is clear from both Figure~\ref{Fig:ConvGraph} and Table~\ref{Table:ConvList} that the cost function converges in progressively fewer iterations as we increase the number of IRLS steps. However, it should be noted that increasing the number of IRLS steps makes each iteration of our optimization routine slower as well. Therefore, a balance has to be struck between the speed of each iteration and the number of iterations required for convergence. In practice, we have found that using 2 IRLS steps per iteration for pairwise registration and 3 IRLS steps per iteration for joint multiview registration yield our desired results. In any event, the key observation is that the single iteration of the FGR is insufficient and yields poorer convergence properties compared to our formulation.\\
\noi\textbf{Choice of Loss Function $\rho(\cdot)$:} The choice of the loss function $\rho(.)$ to be used is a critical factor in our estimation procedure. All loss functions achieve robustness to outliers by a trade-off with statistical efficiency (i.e. accuracy). In practice, the accuracy achievable by a chosen loss function depends on the empirical nature of the error distribution. As discussed earlier, in~\cite{FGR} the authors use the Geman-McClure loss function $\rho(x) = \frac{\mu x^2}{\mu + x^2}$. Here the performance critically depends on chosing a good value for $\mu$ that reflects the outlier distribution inherent to the data used. In~\cite{FGR} the authors start with a large $\mu$ and progressively reduce it in fixed amounts with every iteration. However, if the nature of the data varies significantly, such a fixed annealing procedure may not produce the best possible results. To overcome this limitation of the Geman-McClure method, as we have shown in Section~\ref{Sec:Results}, we have also tested for $L_1$ and $L_{\frac{1}{2}}$ and demonstrated that the latter provides the best performance. Apart from improved performance, an important desirable property of using $L_{\frac{1}{2}}$ is that it is entirely parameter free, hence we do not need to follow any additional annealing procedure. We could conceivably further improve performance by determing the optimal exponent $p$ in $L_{p} (.) = {\left\Vert . \right\Vert}^{p}_{2}$ for $0 < p < 1$. However, for a broad range of possible error distributions, we find that $L_{\frac{1}{2}}$ is adequate.

\subsection{Limitation of FPFH Feature-Matching Based Registration} \label{subsec:FPFH_limitations}
\begin{figure}[t]
    \centering
\includegraphics[width=0.8\linewidth]{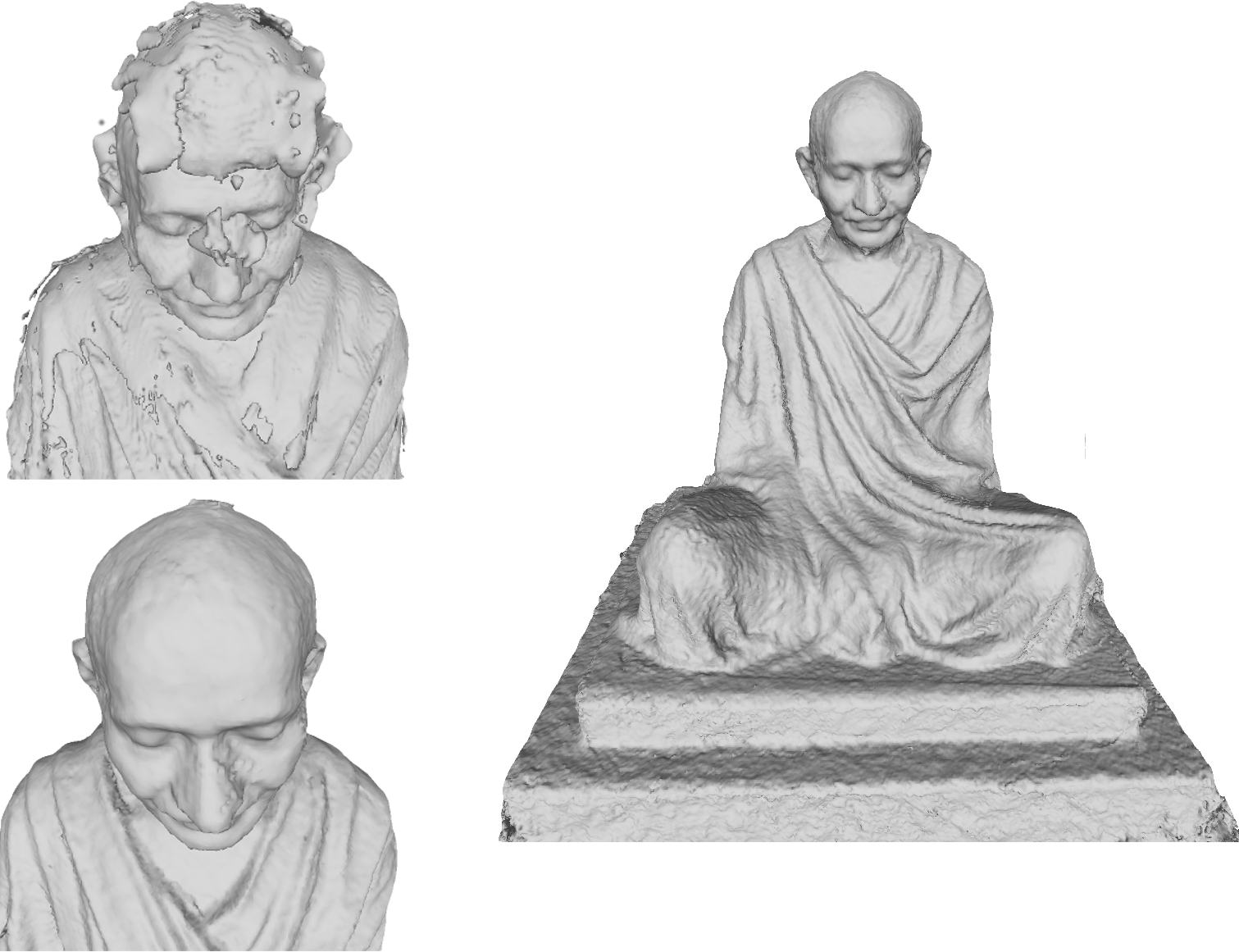}
\caption{3D reconstruction of a statue of Mahatma Gandhi. The close-up on the top left shows that joint multiview registration using FPFH features fails whereas the close-up on the bottom left shows successful registration using our robust pairwise motion estimation within a multiview ICP routine. The full reconstruction is shown on the right.}
\label{Fig:Gandhi}
\end{figure}

In Section~\ref{Sec:Results}, we have demonstrated the potency of the FPFH feature-matching based registration paradigm. However, we note that these experiments, derived from those presented in~\cite{FGR}, have special data characteristics. Specifically, in these examples, either the datasets are small or the motions between adjacent scans are small in magnitude and exhibit temporal smoothness. However, we note here that registration based on feature correspondences cannot work in all possible contexts. For example, when scans in the input dataset have sufficiently large depth differences or large motions between them in both rotation and translation, FPFH feature-based matches become few and unreliable. Consequently, the reconstructions given by both FGR~\cite{FGR} and our proposed algorithms become incorrect. In such a scenario, we need to take recourse to using a robust ICP-based multiview registration method (albeit with a greater computational cost) which converges to the correct solution.\\
\indent In this alternative approach, following~\cite{Govindu2014}, we again consider the camera viewgraph $\mathcal{G} = \{ \mathcal{V}, \mathcal{E} \}$. For each edge available in the set $\mathcal{E}$, we estimate the pairwise motion using a robust version of ICP. Specifically, for the motion estimation step in our robust ICP, we use the motion obtained using the $L_{\frac{1}{2}}$ loss optimization for pairwise registration method as described in Algorithm~\ref{Algo:PairwiseIRLS}. After all pairwise motions represented by the edges $\mathcal{E}$ are estimated, we estimate the absolute motion of each camera w.r.t. a global frame of reference using robust motion averaging. Here, our solution for robust motion averaging for rigid Euclidean motions is similar to the robust rotation averaging method of~\cite{RRAPAMI}. We find that typically, we achieve the desired convergence in 3 full iterations of this procedure. To illustrate our argument, we show the qualities of reconstructions achieved for a life-size statue of Mahatma Gandhi, where the input set of scans is small and mostly have low overlap. Moreover, overlapping scans have significant depth differences between them, leading to significantly different FPFH features and consequently, a high percentage of incorrect matches. A visual representation of this scenario is given in the appendix for better understanding. As we can see in Figure~\ref{Fig:Gandhi}, the joint multiview registration using FPFH features-based matches fails to correctly align some of the scans, whereas the robust multiview-based ICP routine successfully produces a correct reconstruction.

\section{Conclusion}
We have presented a fast and efficient 3D point registration method that optimizes on the $\mathbb{SE}(3)$ group and generalizes the line process based optimization method. In addition, we have shown that in scenarios where feature correspondences cannot be reliably established, our robust motion estimation can make a multiview ICP method robust and effective.

{\small
\bibliographystyle{ieeetr}
\bibliography{iccv_se3_refs}
}

\clearpage

\appendix
\section{Linear Form of the Error Term in Eqn.~\ref{Eqn:LinearModel}}
Let the given set of correspondences be:
\begin{equation*}
\begin{split}
\big\{ (\bfp^s, \bfq^s) \, \lvert \, 1 \leq s \leq \mathcal{S} \, ; \, \bfp^s = \begin{bmatrix} (\bfp^s)^\prime & 1 \end{bmatrix}^\top , \\
\bfq^s = \begin{bmatrix} (\bfq^s)^\prime & 1 \end{bmatrix}^\top \text{ for } \{(\bfp^s)^\prime, (\bfq^s)^\prime\} \in \bbR^3 \big\}.
\end{split}
\end{equation*}
Then we can write the error term in Eqn.~\ref{Eqn:LinearModel} as
\begin{align}
\begin{split}
e^s(\bfM(k)) &= \Bigg\lVert \begin{bmatrix} (\bfp^s)^\prime \\ 1 \end{bmatrix} - \\
& \qquad (\bfI_4 + \Delta\mathfrak{m}(k))\bfM(k-1)\begin{bmatrix} (\bfq^s)^\prime \\ 1 \end{bmatrix} \Bigg\rVert
\end{split} \\
\begin{split}
\Rightarrow e^s(\bfM(k)) &= \Bigg\lVert \begin{bmatrix} (\bfp^s)^\prime-(\bfq^s)^\prime \\ 0 \end{bmatrix} - \\
& \qquad \left[\begin{array}{ccc}[\boldsymbol\omega]_\times&|&\bfu\\\hline\mathbf{0}&|&0\end{array}\right] \bfM(k-1)\begin{bmatrix} (\bfq^s)^\prime \\ 1 \end{bmatrix} \Bigg\rVert
\end{split} \label{Eqn:LinearModelExpanded}
\end{align}
where we get $\Delta\mathfrak{m}(k) = \left[\begin{array}{ccc}[\boldsymbol\omega]_\times&|&\bfu\\\hline\mathbf{0}&|&0\end{array}\right]$ from Eqn.~\ref{Eqn:SE3_se3}. Rewriting Eqn.~\ref{Eqn:LinearModelExpanded} with $\mathfrak{v} = \begin{bmatrix} \boldsymbol\omega & \bfu \end{bmatrix}^\top$, $\bfM(k-1) = \left[\begin{array}{ccc}\bfR(k-1)&|&\bft(k-1)\\\hline\mathbf{0}&|&1\end{array}\right]$ (from Eqn.~\ref{Eqn:SE3MatrixForm}), and by dropping the trailing $0$ for ease of representation, we get,
\begin{equation}
\begin{split}
e^s(\bfM(k)) = \lVert \begin{bmatrix} -\left[\bfR(k-1)(\bfq^s)^\prime+\bft(k-1) \right]_\times \, | \, \bfI_3 \end{bmatrix} \mathfrak{v} - \\
((\bfp^s)^\prime-(\bfq^s)^\prime) \rVert.
\end{split}
\end{equation}
Thus, we have in Eqn.~\ref{Eqn:LinearModel},
\begin{align}
\bfA^s &= \begin{bmatrix} -\left[\bfR(k-1)(\bfq^s)^\prime+\bft(k-1) \right]_\times \,  | \, \bfI_3 \end{bmatrix}, \\
\bfb^s &= (\bfp^s)^\prime-(\bfq^s)^\prime.
\end{align}

\section{Algorithm for Joint Multiview Registration}
Similar to our algorithm for robust motion estimation between a pair of 3D scans, we state our solution for the robust motion estimation of a set of $N (\geq 2)$ 3D scans as given in Algorithm~\ref{Algo:JointMVIRLS}.

\begin{algorithm}[h]
\caption{IRLS estimation of joint multiview 3D registration}\label{Algo:JointMVIRLS}
Input: $\bigg\{\Big\{\big(\bfp_i^1,\bfp_j^1\big) \cdots \big(\bfp_i^{\mathcal{S}_{ij}}, \bfp_j^{\mathcal{S}_{ij}}\big)\Big\} \: \Big\lvert \: (i, j) \in \mathcal{E} \bigg\}$ \big(according to the viewgraph $\mathcal{G} = \{ \mathcal{V}, \mathcal{E} \}$\big) \\
Output: $\bbM = \left\{\bfM_i \: \lvert \: \bfM_i \in \mathbb{SE}(3) \: \forall i \in [1 \cdots N] \right\}$  (Robust estimate of motion of the set of $N = \lvert \mathcal{V} \rvert$ scans) \\
Initialization: $\bbM$ is set to initial guess $\bbM_{initial}$

\begin{algorithmic}[l]
\While {$||\mathbbm{v}|| > N\epsilon$} \Comment{$\mathbbm{v} = \begin{bmatrix} \mathfrak{v}_1 & \cdots & \mathfrak{v}_N \end{bmatrix}^\top $}
\State 1. Compute $\left\{\left(\bfA_{ij}^s,\bfb_{ij}^s \right) \: \big\lvert \: \forall s \in [1 \cdots {\mathcal{S}_{ij}}] \: ; \: (i, j) \in \mathcal{E} \right\}$ using Eqn.~\ref{Eqn:LinearModelMV}
\State 2. Compute weights $w_{ij}^s = \frac{\rho^{\prime}(e_{ij}^s)}{e_{ij}^s}$ as used in Eqn.~\ref{Eqn:LinearSystemMV}
\State 3. Estimate $\mathbbm{v}$ as IRLS solution for Eqn.~\ref{Eqn:LinearSystemMV}
\State 4. Update $\bfM_i \leftarrow \exp(\hat{\mathfrak{v_i}}) \bfM_i$ $\forall i \in [1 \cdots N]$
\EndWhile
\end{algorithmic}
\end{algorithm}

\section{More Results}
\noi\textbf{UWA dataset:} Table~\ref{Table:Traj&TimeUWA} reports the performace of the motion step of the Fast Global Registration (FGR) method of ~\cite{FGR} as well as all the 3 versions of our method on the UWA dataset~\cite{UWAdataset}. This dataset consists of 5 objects and 50 scenes, with each scene consisting of a combination of these objects. The task is to align individual objects to the scenes, given that each scene contains substantial clutter and occlusion. A total of 188 such alignment tasks are provided in the dataset. As we can see from Table~\ref{Table:Traj&TimeUWA}, the $L_{\frac{1}{2}}$ version of our method produces the lowest median rotation angle and median translation norm errors. It is also significantly faster than FGR~\cite{FGR}.\\
\begin{table}[t]
    \centering
    \caption{Median rotation angle error (RAE) (in degrees), median translation norm error (TNE) (in units of the mean scene diameter) and mean running time (in milliseconds) of the motion step of each method for each sequence in the UWA dataset.}
    \label{Table:Traj&TimeUWA}
    \begin{tabular}{| L{1.4cm} || C{1.6cm} | C{1.6cm} | C{1.6cm} | }
        \hline
        Method & Median RAE & Median TNE & Mean Time \\
        \hline
        FGR~\cite{FGR} & 1.276 & 0.152 & 32.7 \\
        \hline
        Our $L_{\frac{1}{2}}$ & \textbf{0.319} & \textbf{0.034} & 10.1 \\
        \hline
        Our $L_1$ & 0.824 & 0.108 & \textbf{7.0} \\
        \hline
        Our GM & 1.276 & 0.152 & 14.5 \\
        \hline
    \end{tabular}
\end{table}

\noi\textbf{Relative improvement of joint multiview approach over two-stage motion averaging approach:} As described in Section 5.2, we show in Table~\ref{Table:RegB&AMV} that the reconstruction error of the scenes in the Augmented ICL-NUIM dataset~\cite{RRIS} decreases when we use our joint multiview estimation procedure on top of the two-stage motion averaged approach. We show the improvement achieved using the $L_{\frac{1}{2}}$ loss, which is our best-performing version.\\
\begin{table}[t]
    \centering
    \caption{Mean registration error (in meters) before and after applying the joint multiview (MV) procedure with $L_{\frac{1}{2}}$ loss on the motion averaged estimate, for full reconstruction from the fragments of each sequence in the Augmented ICL-NUIM dataset.}
    \label{Table:RegB&AMV}
    \begin{tabular}{| l || C{2.7cm} | C{2.7cm} | C{2.7cm} | C{2.7cm} | }
        \hline
        & \texttt{livingroom 1} & \texttt{livingroom 2} & \texttt{office 1} & \texttt{office 2} \\
        \hline
        Before MV & 0.07 & 0.07 & 0.06 & 0.07 \\
        \hline
        After MV & \textbf{0.04} & \textbf{0.05} & \textbf{0.03} & \textbf{0.04} \\
        \hline
    \end{tabular}
\end{table}

\begin{figure}[t]
    \centering
    \includegraphics[width=1\textwidth]{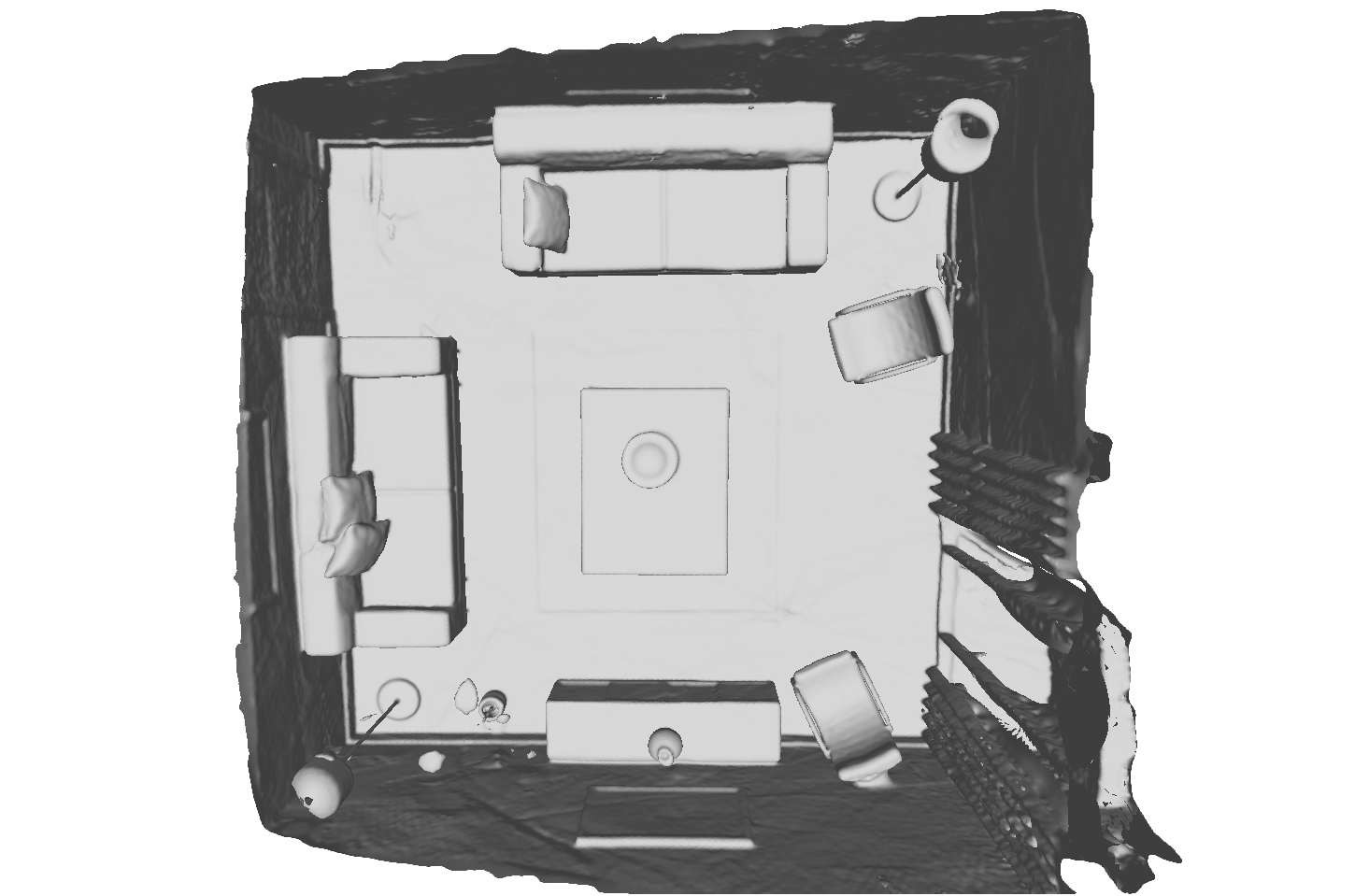}
    \caption{Reconstruction of the \texttt{livingroom 1} sequence from the Augmented ICL-NUIM dataset, as given by our method with the $L_{\frac{1}{2}}$ loss function}
    \label{Fig:Lvr1nRecons}
\end{figure}

\begin{figure}[t]
    \centering
    \includegraphics[width=1\textwidth]{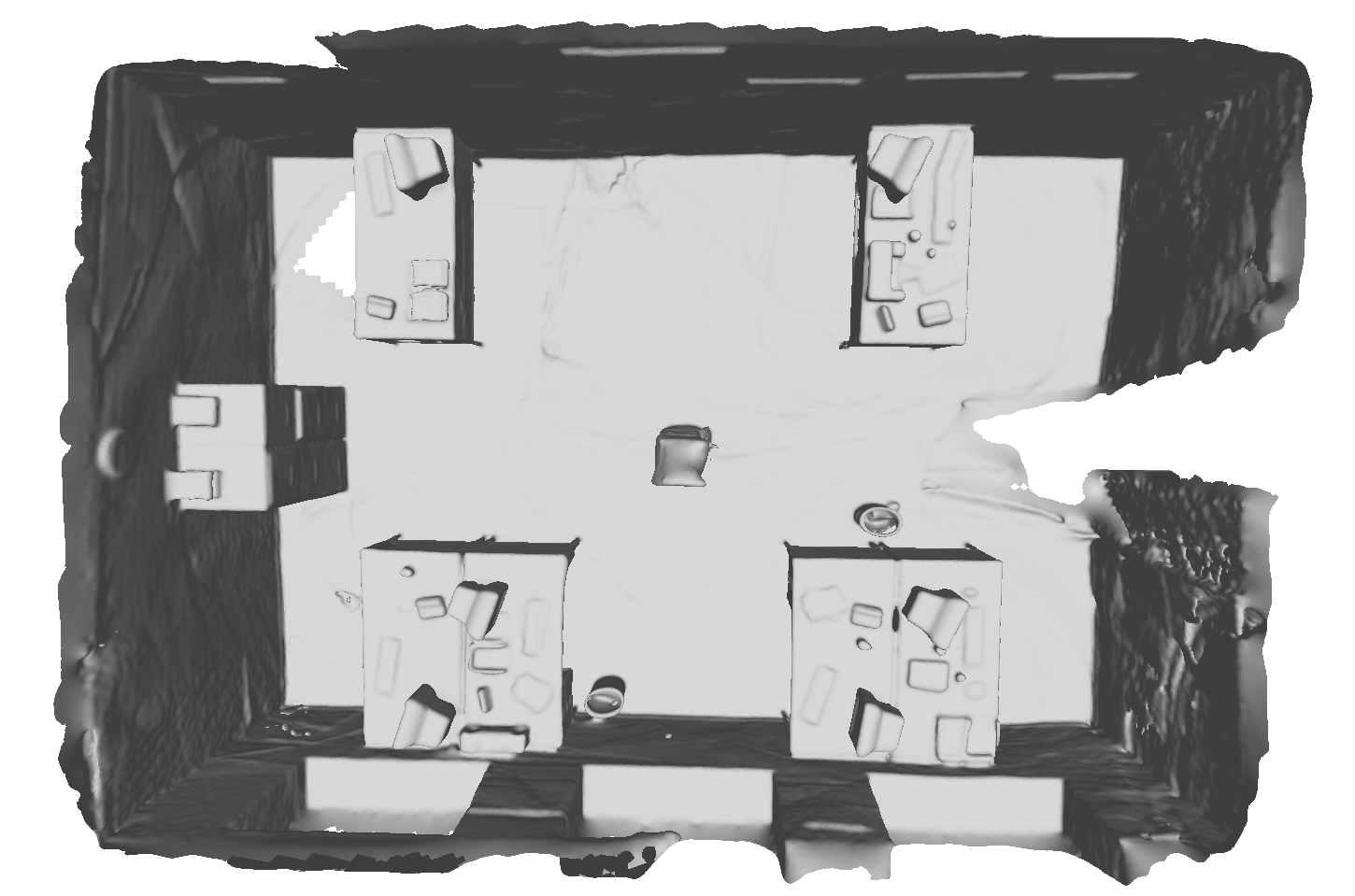}
    \caption{Reconstruction of the \texttt{office 1} sequence from the Augmented ICL-NUIM dataset, as given by our method with the $L_{\frac{1}{2}}$ loss function}
    \label{Fig:Ofc1nRecons}
\end{figure}

\begin{figure}[t]
    \centering
    \includegraphics[width=1\textwidth]{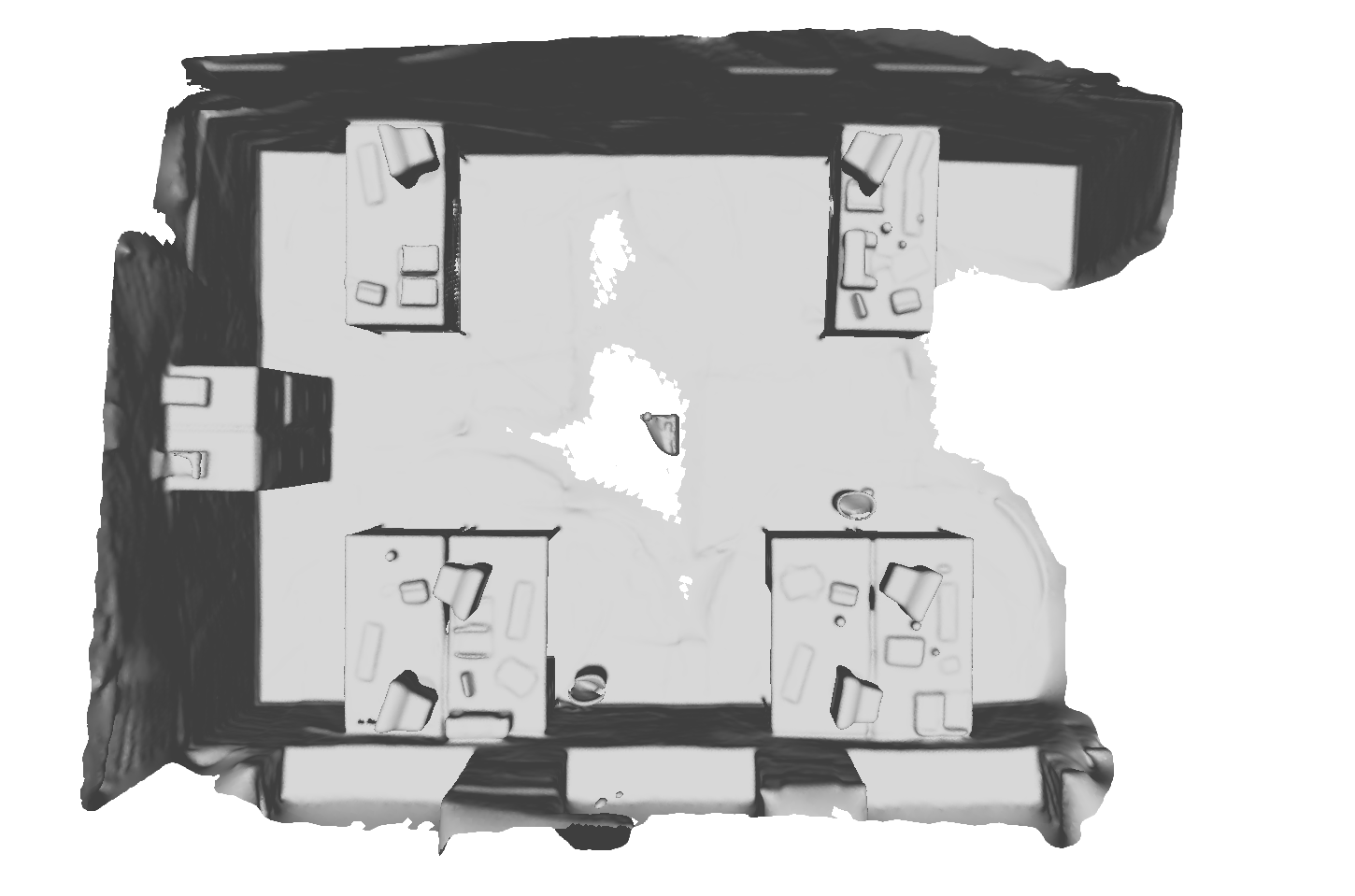}
    \caption{Reconstruction of the \texttt{office 2} sequence from the Augmented ICL-NUIM dataset, as given by our method with the $L_{\frac{1}{2}}$ loss function}
    \label{Fig:Ofc2nRecons}
\end{figure}

\noindent \textbf{Augmented ICL-NUIM dataset:} We also show full reconstructions of the \texttt{livingroom 1}, \texttt{office 1} and \texttt{office 2} sequences from the Augmented ICL-NUIM dataset~\cite{RRIS} in Figures~\ref{Fig:Lvr1nRecons},~\ref{Fig:Ofc1nRecons} and~\ref{Fig:Ofc2nRecons} respectively, as provided by the $L_{\frac{1}{2}}$ version of our method.

\section{An Illustration of the Limitation of FPFH Feature-Matching Based Registration}
\begin{figure}[t]
    \centering
    \includegraphics[width=1\textwidth]{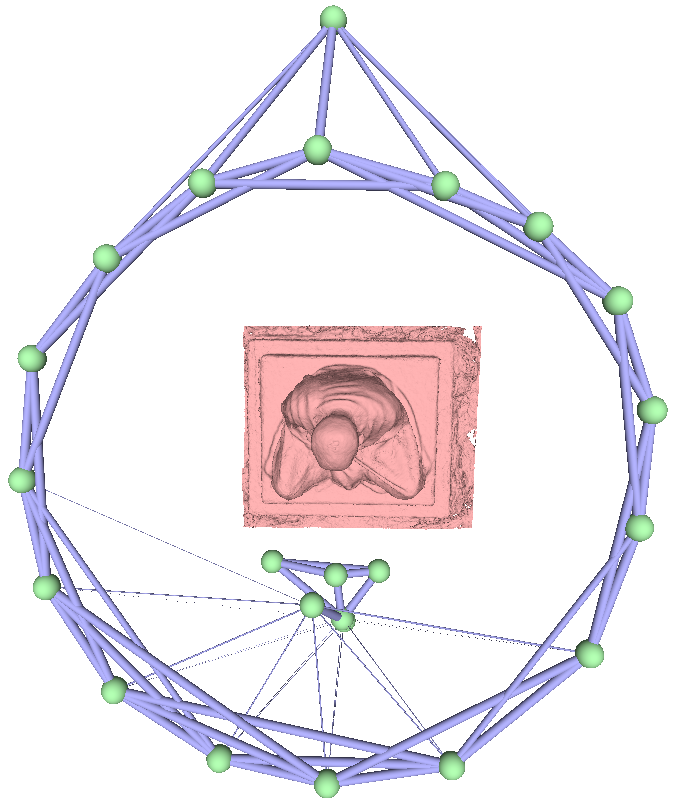}
    \caption{Plan view of a schematic representation of the viewgraph used for reconstruction of the statue of Mahatma Gandhi. See text for details.}
    \label{Fig:GandhiCameras}
\end{figure}

As discussed in Section~\ref{subsec:FPFH_limitations}, we have presented a scenario where the FPFH feature-matching based registration technique breaks down due to unreliability of the feature matches themselves. In this particular scenario, we have 23 scans of a life-size statue of Mahatma Gandhi collected with a standard commercial depth camera. Figure~\ref{Fig:GandhiCameras} shows the plan view of a schematic of the cameras (represented as balls) around the statue, as recovered by our ICP-based multiview approach. Recall that these cameras are the nodes of the viewgraph $\mathcal{G} = \{\mathcal{V}, \mathcal{E}\}$. We also display a schematic of the edges in the viewgraph $\mathcal{G}$ (using sticks). The thickness of each edge is proportional to the number of FPFH feature matches found between the correponding camera (or equivalently scan) pair.\\
\indent We can observe from Figure~\ref{Fig:GandhiCameras} that the thinnest edges are found between pairs of cameras at different depths, implying that there are extremely few FPFH feature matches between these cameras. Compounding this observation with the fact that FPFH features are noisy to begin with, the resultant motions between these cameras, even with our robust cost function, are grossly incorrect. In contrast, our ICP-based multiview approach can, albeit at a higher computational cost, align these cameras correctly and produce the desired reconstruction.

\end{document}

%% file: myincludes.tex
\newcommand{\ba}{\begin{array}}
\newcommand{\ea}{\end{array}}
\newcommand{\noi}{\noindent}

\newcommand{\bc}{\begin{center}}
\newcommand{\ec}{\end{center}}
\newcommand{\bit}{\begin{itemize}}
\newcommand{\eit}{\end{itemize}}

\newcommand{\beq}{\begin{equation}}
\newcommand{\eeq}{\end{equation}}

\newcommand{\bfb}{\mathbf{b}}

\newcommand{\bfp}{\mathbf{p}}
\newcommand{\bfq}{\mathbf{q}}

\newcommand{\bft}{\mathbf{t}}
\newcommand{\bfu}{\mathbf{u}}

\newcommand{\bfw}{\mathbf{w}}

\newcommand{\bfA}{\mathbf{A}}

\newcommand{\bfI}{\mathbf{I}}

\newcommand{\bfL}{\mathbf{L}}
\newcommand{\bfM}{\mathbf{M}}

\newcommand{\bfP}{\mathbf{P}}

\newcommand{\bfR}{\mathbf{R}}

\newcommand{\bfW}{\mathbf{W}}

\newcommand{\bbA}{\mathbb{A}}

\newcommand{\bbM}{\mathbb{M}}

\newcommand{\bbR}{\mathbb{R}}

\newcommand{\bbW}{\mathbb{W}}

\newcommand{\be}{\begin{equation}}
\newcommand{\ee}{\end{equation}}
\newcommand{\beqa}{\begin{eqnarray}}
\newcommand{\eeqa}{\end{eqnarray}}

%% file: iccv_se3_main.bbl
\begin{thebibliography}{10}

\bibitem{KinectFusion}
R.~A. Newcombe, S.~Izadi, O.~Hilliges, D.~Molyneaux, D.~Kim, A.~J. Davison,
  P.~Kohli, J.~Shotton, S.~Hodges, and A.~Fitzgibbon, ``Kinect{F}usion:
  {R}eal-{T}ime {D}ense {S}urface {M}apping and {T}racking,'' in {\em
  Proceedings of the 10th {IEEE} {I}nternational {S}ymposium on {M}ixed and
  {A}ugmented {R}eality}, 2011.

\bibitem{sturm2012benchmark}
J.~Sturm, N.~Engelhard, F.~Endres, W.~Burgard, and D.~Cremers, ``A {B}enchmark
  for the {E}valuation of {RGB-D} {SLAM} {S}ystems,'' in {\em {I}nternational
  {C}onference on {I}ntelligent {R}obots and {S}ystems ({IROS}), {IEEE}/{RSJ}},
  pp.~573--580, {IEEE}, 2012.

\bibitem{Zhang2012}
Z.~Zhang, ``Microsoft {K}inect {S}ensor and its {E}ffect,'' {\em {IEEE}
  {M}ultimedia}, vol.~19, no.~2, pp.~4--10, 2012.

\bibitem{Salvi2007}
J.~Salvi, C.~Matabosch, D.~Fofi, and J.~Forest, ``A {R}eview of {R}ecent
  {R}ange {I}mage {R}egistration {M}ethods with {A}ccuracy {E}valuation,'' {\em
  Image and {V}ision {C}omputing}, vol.~25, no.~5, pp.~578--596, 2007.

\bibitem{Tam2013}
G.~K. Tam, Z.-Q. Cheng, Y.-K. Lai, F.~C. Langbein, Y.~Liu, D.~Marshall, R.~R.
  Martin, X.-F. Sun, and P.~L. Rosin, ``Registration of 3{D} {P}oint {C}louds
  and {M}eshes: {A} {S}urvey from {R}igid to {N}onrigid,'' {\em {IEEE}
  {T}ransactions on {V}isualization and {C}omputer {G}raphics}, vol.~19, no.~7,
  pp.~1199--1217, 2013.

\bibitem{Guo2014}
Y.~Guo, M.~Bennamoun, F.~Sohel, M.~Lu, and J.~Wan, ``3{D} {O}bject
  {R}ecognition in {C}luttered {S}cenes with {L}ocal {S}urface {F}eatures: {A}
  {S}urvey,'' {\em {IEEE} {T}ransactions on {P}attern {A}nalysis and {M}achine
  {I}ntelligence}, vol.~36, no.~11, pp.~2270--2287, 2014.

\bibitem{Mellado2014}
N.~Mellado, D.~Aiger, and N.~J. Mitra, ``Super 4{PCS} {F}ast {G}lobal
  {P}ointcloud {R}egistration via {S}mart {I}ndexing,'' in {\em Computer
  {G}raphics {F}orum}, vol.~33 (5), pp.~205--215, Wiley {O}nline {L}ibrary,
  2014.

\bibitem{Gelfand2005}
N.~Gelfand, N.~J. Mitra, L.~J. Guibas, and H.~Pottmann, ``Robust {G}lobal
  {R}egistration,'' in {\em Symposium on {G}eometry {P}rocessing}, vol.~2 (3),
  p.~5, 2005.

\bibitem{Makadia2006}
A.~Makadia, A.~Patterson, and K.~Danilidis, ``Fully {A}utomatic {R}egistration
  of 3{D} point {C}louds,'' in {\em {IEEE} {C}omputer {S}ociety {C}onference on
  {C}omputer {V}ision and {P}attern {R}ecognition}, vol.~1, pp.~1297--1304,
  {IEEE}, 2006.

\bibitem{Holz2015}
D.~Holz, A.~E. Ichim, F.~Tombari, R.~B. Rusu, and S.~Behnke, ``Registration
  with the {P}oint {C}loud {L}ibrary: {A} {M}odular {F}ramework for {A}ligning
  in 3{D},'' {\em {IEEE} {R}obotics \& {A}utomation {M}agazine}, vol.~22,
  no.~4, pp.~110--124, 2015.

\bibitem{RRIS}
S.~Choi, Q.-Y. Zhou, and V.~Koltun, ``Robust {R}econstruction of {I}ndoor
  {S}cenes,'' in {\em {IEEE} {C}onference on {C}omputer {V}ision and {P}attern
  {R}ecognition}, pp.~5556--5565, {IEEE}, 2015.

\bibitem{FGR}
Q.-Y. Zhou, J.~Park, and V.~Koltun, ``Fast {G}lobal {R}egistration,'' in {\em
  European {C}onference on {C}omputer {V}ision}, pp.~766--782, Springer, 2016.

\bibitem{Tombari2013}
F.~Tombari, S.~Salti, and L.~Di~Stefano, ``Performance {E}valuation of 3{D}
  {K}eypoint {D}etectors,'' {\em International {J}ournal of {C}omputer
  {V}ision}, vol.~102, no.~1-3, pp.~198--220, 2013.

\bibitem{Guo2016}
Y.~Guo, M.~Bennamoun, F.~Sohel, M.~Lu, J.~Wan, and N.~M. Kwok, ``A
  {C}omprehensive {P}erformance {E}valuation of 3{D} {L}ocal {F}eature
  {D}escriptors,'' {\em International {J}ournal of {C}omputer {V}ision},
  vol.~116, no.~1, pp.~66--89, 2016.

\bibitem{Rusinkiewicz2001}
S.~Rusinkiewicz and M.~Levoy, ``Efficient {V}ariants of the {ICP}
  {A}lgorithm,'' in {\em {P}roceedings of the {T}hird {I}nternational
  {C}onference on 3{D} {D}igital {I}maging and {M}odeling}, pp.~145--152,
  {IEEE}, 2001.

\bibitem{Aiger2008}
D.~Aiger, N.~J. Mitra, and D.~Cohen-Or, ``4-{P}oints {C}ongruent {S}ets for
  {R}obust {P}airwise {S}urface {R}egistration,'' in {\em {ACM} {T}ransactions
  on {G}raphics ({T}o{G})}, vol.~27 (3), p.~85, {ACM}, 2008.

\bibitem{Drost2010}
B.~Drost, M.~Ulrich, N.~Navab, and S.~Ilic, ``Model {G}lobally, {M}atch
  {L}ocally: {E}fficient and {R}obust 3{D} {O}bject {R}ecognition,'' in {\em
  {IEEE} {C}onference on {C}omputer {V}ision and {P}attern {R}ecognition},
  pp.~998--1005, {IEEE}, 2010.

\bibitem{Bouaziz}
S.~Bouaziz, A.~Tagliasacchi, and M.~Pauly, ``Sparse {I}terative {C}losest
  {P}oint,'' in {\em Proceedings of the {E}leventh {E}urographics/{ACM}
  {SIGGRAPH} {S}ymposium on {G}eometry {P}rocessing}, pp.~113--123,
  Eurographics Association, 2013.

\bibitem{chetverikov2002trimmed}
D.~Chetverikov, D.~Svirko, D.~Stepanov, and P.~Krsek, ``The {T}rimmed
  {I}terative {C}losest {P}oint {A}lgorithm,'' in {\em Proceedings of the 16th
  {I}nternational {C}onference on {P}attern {R}ecognition}, vol.~3,
  pp.~545--548, {IEEE}, 2002.

\bibitem{yang2013go}
J.~Yang, H.~Li, and Y.~Jia, ``Go-{ICP}: {S}olving 3{D} {R}egistration
  {E}fficiently and {G}lobally {O}ptimally,'' in {\em {IEEE} {I}nternational
  {C}onference on {C}omputer {V}ision}, pp.~1457--1464, {IEEE}, 2013.

\bibitem{Umeyama1991}
S.~Umeyama, ``Least-{S}quares {E}stimation of {T}ransformation {P}arameters
  {B}etween {T}wo {P}oint {P}atterns,'' {\em {IEEE} {T}ransactions on {P}attern
  Analysis and {M}achine {I}ntelligence}, vol.~13, no.~4, pp.~376--380, 1991.

\bibitem{Raguram2008}
R.~Raguram, J.-M. Frahm, and M.~Pollefeys, ``A {C}omparative {A}nalysis of
  {RANSAC} {T}echniques {L}eading to {A}daptive {R}eal-{T}ime {R}andom {S}ample
  {C}onsensus,'' in {\em European {C}onference on {C}omputer {V}ision},
  pp.~500--513, Springer, 2008.

\bibitem{FPFH}
R.~B. Rusu, N.~Blodow, and M.~Beetz, ``Fast {P}oint {F}eature {H}istograms
  ({FPFH}) for 3{D} {R}egistration,'' in {\em {IEEE} {I}nternational
  {C}onference on {R}obotics and {A}utomation}, pp.~3212--3217, {IEEE}, 2009.

\bibitem{Papazov2012}
C.~Papazov, S.~Haddadin, S.~Parusel, K.~Krieger, and D.~Burschka, ``Rigid 3{D}
  {G}eometry {M}atching for {G}rasping of {K}nown {O}bjects in {C}luttered
  {S}cenes,'' {\em The {I}nternational {J}ournal of {R}obotics {R}esearch},
  vol.~31, no.~4, pp.~538--553, 2012.

\bibitem{Salas2013}
R.~F. Salas-Moreno, R.~A. Newcombe, H.~Strasdat, P.~H. Kelly, and A.~J.
  Davison, ``{SLAM}++: {S}imultaneous {L}ocalisation and {M}apping at the
  {L}evel of {O}bjects,'' in {\em {IEEE} {C}onference on {C}omputer {V}ision
  and {P}attern {R}ecognition}, pp.~1352--1359, {IEEE}, 2013.

\bibitem{Hartley2007}
R.~I. Hartley and F.~Kahl, ``Global {O}ptimization {T}hrough {S}earching
  {R}otation {S}pace and {O}ptimal {E}stimation of the {E}ssential {M}atrix,''
  in {\em {IEEE} 11th {I}nternational {C}onference on {C}omputer {V}ision},
  pp.~1--8, {IEEE}, 2007.

\bibitem{Li2007}
H.~Li and R.~Hartley, ``The 3{D}-3{D} {R}egistration {P}roblem {R}evisited,''
  in {\em {IEEE} 11th {I}nternational {C}onference on {C}omputer {V}ision},
  pp.~1--8, {IEEE}, 2007.

\bibitem{Enqvist2009}
O.~Enqvist, K.~Josephson, and F.~Kahl, ``Optimal {C}orrespondences from
  {P}airwise {C}onstraints,'' in {\em {IEEE} 12th {I}nternational {C}onference
  on {C}omputer {V}ision}, pp.~1295--1302, {IEEE}, 2009.

\bibitem{Yang2016}
J.~Yang, H.~Li, D.~Campbell, and Y.~Jia, ``Go-{ICP}: a {G}lobally {O}ptimal
  {S}olution to 3{D} {ICP} {P}oint-{S}et {R}egistration,'' {\em {IEEE}
  {T}ransactions on {P}attern {A}nalysis and {M}achine {I}ntelligence},
  vol.~38, no.~11, pp.~2241--2254, 2016.

\bibitem{JianVemuri}
B.~Jian and B.~C. Vemuri, ``Robust {P}oint {S}et {R}egistration {U}sing
  {G}aussian {M}ixture {M}odels,'' {\em IEEE Transactions on Pattern Analysis
  and Machine Intelligence}, vol.~33, pp.~1633--1645, Aug 2011.

\bibitem{EvangelidisHoraud}
G.~D. Evangelidis and R.~Horaud, ``Joint {A}lignment of {M}ultiple {P}oint
  {S}ets with {B}atch and {I}ncremental {E}xpectation-{M}aximization,'' {\em
  IEEE Transactions on Pattern Analysis and Machine Intelligence}, vol.~40,
  pp.~1397--1410, June 2018.

\bibitem{EckartEtAl}
B.~Eckart, K.~Kim, and J.~Kautz, ``{HGMR}: {H}ierarchical {G}aussian {M}ixtures
  for {A}daptive 3{D} {R}egistration,'' in {\em The European Conference on
  Computer Vision (ECCV)}, September 2018.

\bibitem{Govindu2014}
V.~M. Govindu and A.~Pooja, ``On {A}veraging {M}ultiview {R}elations for 3{D}
  {S}can {R}egistration,'' {\em {IEEE} {T}ransactions on {I}mage {P}rocessing},
  vol.~23, pp.~1289--1302, March 2014.

\bibitem{Torsello11multi-viewregistration}
A.~Torsello and A.~Albarelli, ``Multi-{V}iew {R}egistration via {G}raph
  {D}iffusion of {D}ual {Q}uaternions,'' in {\em {IEEE} {C}onference on
  {C}omputer {V}ision and {P}attern {R}ecognition}, pp.~2441--2448, 2011.

\bibitem{Arrigoni2016}
F.~Arrigoni, B.~Rossi, and A.~Fusiello, ``Global {R}egistration of 3{D} {P}oint
  {S}ets via {LRS} {D}ecomposition,'' in {\em {E}uropean {C}onference on
  {C}omputer {V}ision}, pp.~489--504, 2016.

\bibitem{IRLSHollandWelsch}
P.~W. Holland and R.~E. Welsch, ``Robust {R}egression {U}sing {I}teratively
  {R}eweighted {L}east-{S}quares,'' {\em Communications in {S}tatistics -
  {T}heory and {M}ethods}, vol.~6, no.~9, pp.~813--827, 1977.

\bibitem{Aftab2015}
K.~Aftab and R.~Hartley, ``Convergence of {I}teratively {R}e-weighted {L}east
  {S}quares to {R}obust {M}-estimators,'' in {\em {A}pplications of {C}omputer
  {V}ision (WACV), 2015 {IEEE} {W}inter {C}onference on}, pp.~480--487, IEEE,
  2015.

\bibitem{black1996unification}
M.~J. Black and A.~Rangarajan, ``On the {U}nification of {L}ine {P}rocesses,
  {O}utlier {R}ejection, and {R}obust {S}tatistics with {A}pplications in
  {E}arly {V}ision,'' {\em International {J}ournal of {C}omputer {V}ision},
  vol.~19, no.~1, pp.~57--91, 1996.

\bibitem{Chirikjian02}
G.~Chirikjian, {\em Stochastic {M}odels, {I}nformation Theory, and {L}ie
  {G}roups, {V}olume 2: {A}nalytic {M}ethods and {M}odern {A}pplications}.
\newblock Birkhauser {B}oston, 2011.

\bibitem{SeligRobotics}
J.~Selig, {\em Geometrical {M}ethods in {R}obotics}.
\newblock Springer {N}ew {Y}ork, 2013.

\bibitem{park1995distance}
F.~C. Park, ``Distance {M}etrics on the {R}igid-{B}ody {M}otions with
  {A}pplications to {M}echanism {D}esign,'' {\em Journal of {M}echanical
  {D}esign}, vol.~117, no.~1, pp.~48--54, 1995.

\bibitem{meer1991robust}
P.~Meer, D.~Mintz, A.~Rosenfeld, and D.~Y. Kim, ``Robust {R}egression {M}ethods
  for {C}omputer {V}ision: {A} {R}eview,'' {\em International {J}ournal of
  {C}omputer {V}ision}, vol.~6, no.~1, pp.~59--70, 1991.

\bibitem{ZacurLeftInvariant}
E.~Zacur, M.~Bossa, and S.~Olmos, ``Left-{I}nvariant {R}iemannian {G}eodesics
  on {S}patial {T}ransformation {G}roups,'' {\em {SIAM} {J}ournal on {I}maging
  {S}ciences}, vol.~7, no.~3, pp.~1503--1557, 2014.

\bibitem{SzeliskiBook}
R.~Szeliski, {\em Computer {V}ision: {A}lgorithms and {A}pplications}.
\newblock Springer-{V}erlag {N}ew York, {I}nc., 2010.

\bibitem{DARE}
F.~J. Lawin, M.~Danelljan, F.~Khan, P.-E. Forss\'en, and M.~Felsberg, ``Density
  {A}daptive {P}oint {S}et {R}egistration,'' in {\em {IEEE} Conference on
  Computer Vision and Pattern Recognition}, Computer Vision Foundation, June
  2018.

\bibitem{RRAPAMI}
A.~Chatterjee and V.~M. Govindu, ``Robust {R}elative {R}otation {A}veraging,''
  {\em {IEEE} {T}ransactions on {P}attern {A}nalysis and {M}achine
  {I}ntelligence}, vol.~40, pp.~958--972, April 2018.

\bibitem{Bartoli2013}
A.~Bartoli, D.~Pizarro, and M.~Loog, ``Stratified {G}eneralized {P}rocrustes
  {A}nalysis,'' {\em International journal of computer vision}, vol.~101,
  no.~2, pp.~227--253, 2013.

\bibitem{Bergstrom2014}
P.~Bergstr{\"o}m and O.~Edlund, ``Robust {R}egistration of {P}oint {S}ets
  {U}sing {I}teratively {R}eweighted {L}east {S}quares,'' {\em Computational
  Optimization and Applications}, vol.~58, no.~3, pp.~543--561, 2014.

\bibitem{Fitzgibbon2003}
A.~W. Fitzgibbon, ``Robust {R}egistration of 2{D} and 3{D} {P}oint {S}ets,''
  {\em Image and {V}ision {C}omputing}, vol.~21, no.~13-14, pp.~1145--1153,
  2003.

\bibitem{UWAdataset}
A.~S. Mian, M.~Bennamoun, and R.~Owens, ``Three-{D}imensional {M}odel-{B}ased
  {O}bject {R}ecognition and {S}egmentation in {C}luttered {S}cenes,'' {\em
  IEEE {T}ransactions on {P}attern {A}nalysis and {M}achine {I}ntelligence},
  vol.~28, no.~10, pp.~1584--1601, 2006.

\end{thebibliography}
